%% file: acl_lualatex.tex
\documentclass[11pt]{article}

\usepackage[final]{acl}

\usepackage{textcomp}
\usepackage{times}
\usepackage{latexsym}
\usepackage{amsmath}
\usepackage{amssymb}
\usepackage{mathtools}
\usepackage{amsthm}
\usepackage{booktabs}
\usepackage[most]{tcolorbox}
\usepackage{enumitem}

\newtcolorbox{exbox}[1]{breakable, title=#1}
\newtheorem{theorem}{Theorem}[section]

\newtheorem{definition}[theorem]{Definition}

\theoremstyle{remark}

\usepackage{rotating}
\usepackage{bm}
\usepackage{array}
\usepackage{tikz}
\usepackage{multirow}
\usepackage{xcolor}  
\usepackage[table]{xcolor}
\usepackage[T1]{fontenc}

\usepackage[utf8]{inputenc}

\usepackage{microtype}

\usepackage{inconsolata}
\usepackage{wrapfig}
\usepackage{subcaption}
\usepackage{graphicx}
\newcommand{\model}{\bm{\theta}}
\newcommand{\sample}{\bm{\xi}}
\newcommand{\method}{\textbf{\textit{LearnAlign}}}
\usepackage[bottom]{footmisc}
\usepackage{hyperref}

\usepackage[english,bidi=default]{babel} 
\babelfont{rm}{TeXGyreTermesX} 
\babelprovide[import]{hindi}
\babelfont[*devanagari]{rm}{Lohit Devanagari}
\babelprovide[import]{arabic}
\babelfont[*arabic]{rm}{Noto Sans Arabic}

\usepackage{microtype}
\microtypesetup{protrusion=true, expansion=false}

\title{\textit{LearnAlign}: Data Selection for LLM Reinforcement Learning with Improved Gradient Alignment}

\author{
  \textbf{Shipeng Li\textsuperscript{1}\thanks{Equal contribution}},
  \textbf{Zhiqin Yang\textsuperscript{3}$^*$},
  \textbf{Shikun Li\textsuperscript{2}$^*$\thanks{Corresponding authors}},
    \textbf{Xiaobo Xia\textsuperscript{4}},
  \textbf{Hengyu Liu\textsuperscript{3}},
  \textbf{Xinghua Zhang\textsuperscript{5}},
\\
  \textbf{Gaode Chen\textsuperscript{5}},
   \textbf{Dong Fang\textsuperscript{6}}, 
 \textbf{Ying Tai\textsuperscript{1}},
  \textbf{Zhe Peng\textsuperscript{2}$^{\dagger}$}
\\
\\
\textsuperscript{1} Nanjing University, 
 \textsuperscript{2} The Hong Kong Polytechnic University, \\
  \textsuperscript{3} The Chinese University of Hong Kong, 
  \textsuperscript{4} University of Science and Technology of China, \\
  \textsuperscript{5} Institute of Information Engineering, Chinese Academy of Sciences, 
  \textsuperscript{6} LIGHTSPEED
\\{
  \href{mailto:shipengli.nju@gmail.com}{shipengli.nju@gmail.com}, \href{mailto:shikunli.ml@gmail.com}{shikunli.ml@gmail.com}, \href{mailto:jeffrey-zhe.peng@polyu.edu.hk}{jeffrey-zhe.peng@polyu.edu.hk}
  }
}

\begin{document}

\maketitle
\begin{abstract}
Reinforcement learning with verifiable rewards (RLVR) has become a key technique for enhancing LLMs' reasoning abilities, yet its data inefficiency remains a major bottleneck. To address this critical yet challenging issue, we present a novel gradient-alignment-based method, named~\textit{LearnAlign}, which intelligently selects the learnable and representative training reasoning data for RLVR post-training. To overcome the well-known response-length bias in gradient norms, we introduce the data learnability based on the success rate, which indicates the learning potential of each data point.
Experiments across five reasoning benchmarks show that our method significantly reduces training data requirements while achieving minor performance degradation or even improving performance compared to full-data training. Specifically, it reduces data requirements by up to 1,000 data points with better performance (77.5$\%$) than that on the full dataset on the GSM8K benchmark (77.0$\%$). Furthermore, its efficiency is demonstrated on both mathematical and code benchmarks by using much less data from the DAPO-MATH-17K dataset.
\end{abstract}

\input{main/intro}
\input{main/related}
\input{main/preliminary}
\input{main/method}
\input{main/exp}
\section{Conclusion}
In this study, we propose a novel data selection framework for RLVR post‑training of LLMs, driven by a gradient‑alignment method. Building upon policy-gradient direction alignment, our framework introduces a success-rate-based learnability score to mitigate response-length bias and efficiently identify a compact subset of reasoning examples. Experiments on five benchmarks demonstrate that, with only approximately 1,000 samples (less than 15\% of the full training set), our method matches or surpasses the performance of full‑data training on both in-distribution and out-of-distribution tasks.
\section*{Limitations} Due to limited GPU resources, we only evaluate the effectiveness of data selection methods on relatively small-scale models (1.5B, 3B, and 7B models) and datasets. Specifically, our current assessment of the proposed method’s effectiveness focuses on math reasoning datasets, including GSM8K and DAPO-MATH-17k. In the future, we plan to evaluate it
 on larger models and diverse datasets.
We believe this work establishes an effective paradigm for data‑efficient RLVR fine‑tuning. Future research directions may encompass the extension to a broader range of task domains, the integration of dynamic curricula with adaptive selection strategies, and the pursuit of alignment with out-of-distribution data.
\bibliography{custom}

\newpage
\appendix

\section{Additional Experimental Details}

\subsection{Hyperparameters and prompt}
\label{Additional Experimental Details}
For additional experimental hyperparameters, please refer to Table~\ref{tab:app_1}. The prompts used for GSM8K and DAPO-MATH-17K are as follows:

\begin{exbox}{The System Prompt for GSM8K:}
A conversation between User and Assistant. The user asks a question, and the Assistant solves it. The assistant first thinks about the reasoning process in the mind and then provides the user with the answer. The reasoning process and answer are enclosed within <think> </think> and <answer> </answer> tags, respectively, i.e., <think> reasoning process here </think> <answer> answer here </answer>.
\end{exbox}

\begin{exbox}{The System Prompt for DAPO-MATH-17K:}
Let's think step by step and output the final answer within \textbackslash boxed\{\}.
\end{exbox}

\begin{table*}[ht]
\caption{More detailed experimental parameter setting.}
\begin{center}
\resizebox{0.9\textwidth}{!}{
\begin{tabular}{lcc}
\toprule[1.5pt]
\textbf{Training Dataset} & \textbf{GSM8K} & \textbf{DAPO-MATH-17K} \\
\midrule[0.6pt]
\multicolumn{3}{l}{\textbf{Training Configuration}} \\
\midrule[0.6pt]
Train Batch Size          & 48    & 64    \\
Max Prompt Length         & 512   & 512   \\
Max Response Length       & 1024  & 2048  \\
Train epochs               & 2 & 2 \\
Clip Ratio                & 0.2   & 0.2   \\
\midrule[0.6pt]
\multicolumn{3}{l}{\textbf{Optimizer Parameters}} \\
\midrule[0.6pt]
Optimizer                 & AdamW ($\beta_1=0.9$, $\beta_2=0.999$, $\epsilon=10^{-8}$) & AdamW ($\beta_1=0.9$, $\beta_2=0.999$, $\epsilon=10^{-8}$) \\
Learning Rate             & 1e-06 & 1e-06 \\
Warmup Style              & Cosine & Cosine \\
Warmup Steps Ratio        & 0.1   & 0.1   \\
KL Loss Coefficient       & 0.04  & 0.04  \\
\midrule[0.6pt]
\multicolumn{3}{l}{\textbf{Temperature}} \\
\midrule[0.6pt]
Training Temperature      & 1.0   & 1.0   \\
Evaluation Temperature    & 0     & 0.8   \\
\bottomrule[1.5pt]
\end{tabular}}
\label{tab:app_1}
\end{center}
\end{table*}

\subsection{Detailed Compared Methods}
\label{appendix:detailed_baselines}
In this section, we detail the baseline methods compared with \textit{LearnAlign}. \textbf{ Random Sampling:} We randomly select a portion of all the datasets as the training set data. \textbf{PPL-Top}~\citep{laurenccon2022bigscience} and
\textbf{PPL-Middle}~\citep{ankner2024perplexed} 
all based on the perplexity calculated by Eq.(\ref{eq:ppl}):
\begin{align}
\label{eq:ppl}
\text{PPL}(\sample) = \exp\left( -\frac{1}{T} \sum_{t=1}^{T} \log \pi_{\model}(y_t | \sample_{0:t-1}) \right),
\end{align}
where PPL-Top selects data with the top perplexity, while PPL-Middle selects the data with the middle perplexity. Furthermore, \textbf{Instruction-Following Difficulty (IFD)}~\citep{li2023quantity} quantifies the inherent difficulty of an instruction-answer pair for a large language model (LLM). It is calculated as the ratio between the direct answer score $s_{\model}(o)$ and the conditioned answer score $s_{\model}(o|q)$. Direct answer score $s_{\model}(o)$ is the averaged cross-entropy loss of generating the answer $o$ without any instructional context. At the same time, conditioned answer score $s_{\model}(o|q)$ is the averaged cross-entropy loss of generating the ground-truth answer $o$ given the instruction $q$. The IFD is then calculated as:
\begin{align}
\text{IFD}_{\model}=\frac{s_{\model}(o|q)}{s_{\model}(o)},
\end{align}
where a higher IFD score indicates that the instruction provides less benefit to the response generation.
\textbf{Token Length}~\citep{xia2024rethinking} quantifies the value of a sample based on its token count. We calculate the total token length by combining the tokens from both the question and the answer.
\textbf{SelectIT}~\citep{liu2024selectit} harnesses the inherent uncertainty within the LLMs. This approach utilizes a multi-granularity self-reflection mechanism, seamlessly integrating token-level, sentence-level, and parameter-weighted model-level uncertainty analyses to evaluate and rank the quality of instruction data.
\textbf{LIMR} ~\citep{li2025limr} measures the learning impact of each training sample by its alignment with the overall learning trajectory of the model.

\section{Theoretical Motivation for the Learnability Metric}
\label{Justification}
Although the proposed learnability metric $p(1-p)$ may appear simple, it is in fact a theoretically grounded formulation for modeling learnability under Bernoulli feedback in RLVR.

First, the success rate $p$ measures how often the model receives informative positive trajectories revealing correct behavior, while $1-p$ captures the remaining room for improvement. A sample provides a useful learning signal only when both conditions co-exist, and thus their product $p(1-p)$ can represent the expected improvement that a data point will provide~\citep{florensa2018automatic,tzannetos2023proximal}.

Second, $p(1-p)$ is precisely the variance of Bernoulli accuracy rewards. Recent theoretical analyses~\citep{razin2025makes,bae2025online} show that the reward variance lower-bounds the KL divergence between the initial and the optimal model, making it an effective statistical quantity reflecting the gradient informativeness of a sample.

Third, this quadratic form is not arbitrary: it is the unique smooth, symmetric, unimodal function that (i) peaks at intermediate difficulty, (ii) vanishes at $p=1$, and (iii) aligns with Fisher information based measures of sample utility~\citep{mackay1992information}. Alternative function choices fail to satisfy these properties or lack comparable theoretical interpretability.

Last but not least, as shown in Appendix~\ref{theorem}, for a fixed query \(\xi\) and model \(\theta\), the gradient magnitude is positively proportional to \(p(1-p)\). It indicates that \(p(1-p)\)
 can represent the information about the gradient magnitude without the issue of response-length bias.

Overall,  $p(1-p)$ is a principled, theoretically grounded, and empirically supported metric for modeling learnability.

\section{Theoretical Analysis of the Learnability–Gradient Relationship}
\label{theorem}
Here, we prove a theorem to show the gradient magnitude is positively proportional to $p(1-p)$.
Given a prompt $\xi$ and a response $y \in \mathcal{Y}$. 
The policy $\pi_\theta(y \mid \xi)$ has logits $z_{\xi,y}(\theta)$, 
which are functions of the parameters $\theta$; for simplicity, we denote 
them as $z_{\xi,y}$ in the following. Under this notation, the policy satisfies
\[
\pi_{\theta}(y \mid \xi)
= \frac{\exp\!\left(z_{\xi,y}\right)}
       {\sum_{y' \in \mathcal{Y}} \exp\!\left(z_{\xi,y'}\right)}.
\]

Let \(y^*\) be the unique correct action with success probability \(p:=\pi_\theta(y^*\mid\xi)\), binary reward \(r(y)=\mathbf{1}[y=y^*]\), and baseline \(b(\xi) := \mathbb{E}_{y\sim\pi_\theta(\cdot\mid\xi)}[r(y)] = p\).
For simplicity, assume a single incorrect action \(\bar{y}\neq y^*\), with probability \(1-p=\pi_\theta(\bar{y}\mid\xi)\).

\begin{theorem}[Gradient Magnitude Factorization]
For the one-correct-answer setting with binary reward, the policy gradient for a sample \(\xi\) can be written as
\begin{align*}
    \nabla_\theta \mathcal{J}(\theta;\xi)
    = p(1-p)\,\mathbf{d}(\xi,\theta),
\end{align*}
for the direction vector \(\mathbf{d}(\xi,\theta)\in\mathbb{R}^{\dim(\theta)}\).
Consequently, for fixed \(\xi\) and \(\theta\),
\begin{align*}
    \|\nabla_\theta\mathcal{J}(\theta;\xi)\|
    \propto p(1-p),
\end{align*}
i.e., the gradient magnitude is positively proportional to \(p(1-p)\).
\end{theorem}

\begin{proof}
The advantage for an action can be expressed as:
\begin{align*}
    A(\xi,y)=r(y)-b(\xi), \qquad b(\xi)=p,
\end{align*}
where the baseline is chosen to be the constant $b(\xi) = p$. Consequently, 
\begin{align*}
    A(\xi,y^*)=1-p,
    \qquad
    A(\xi,\bar{y})=-p.
\end{align*}
Consider the expected advantage under the policy $\pi_\theta$:
\begin{align*}
  \mathcal{J}(\theta;\xi)
  := \mathbb{E}_{y\sim\pi_\theta(\cdot\mid\xi)}[A(\xi,y)]
\end{align*}
The policy gradient is then given by the standard identity:
\begin{align*}
  \nabla_\theta \mathcal{J}(\theta;\xi)
  =
  \mathbb{E}_{y\sim\pi_\theta(\cdot\mid\xi)}
  \bigl[A(\xi,y)\,\nabla_\theta\log\pi_\theta(y\mid\xi)\bigr].
\end{align*}

For a softmax policy parameterized by logits $z_{\xi,y}$, the score function satisfies:
\begin{align*}
  \frac{\partial\log\pi_\theta(y'\mid\xi)}{\partial z_{\xi,y}}
  = \mathbf{1}[y'=y] - \pi_\theta(y\mid\xi).
\end{align*}
Differentiating $\mathcal{J}(\theta;\xi)$ with respect to a specific logit $z_{\xi,y}$ therefore yields:
\begin{align*}
\frac{\partial \mathcal{J}(\theta;\xi)}{\partial z_{\xi,y}}
&=
\pi_\theta(y\mid\xi)A(\xi,y)\\
&- \pi_\theta(y\mid\xi)\,\mathbb{E}_{y'\sim\pi_\theta}[A(\xi,y')].
\end{align*}

Since the baseline is the expected reward probability,
\begin{align*}
\mathbb{E}_{y'\sim\pi_\theta}[A(\xi,y')]
= \mathbb{E}[r(y') - p] = p - p = 0,
\end{align*}
we obtain the simplified logit gradient:
\begin{align*}
    \frac{\partial \mathcal{J}(\theta;\xi)}{\partial z_{\xi,y}}
    = \pi_\theta(y\mid\xi)A(\xi,y).
\end{align*}
The corresponding logit update is:
\begin{align*}
  \Delta z_{\xi,y} \propto \pi_\theta(y\mid\xi)A(\xi,y),
\end{align*}
Substituting the two possible actions. Let $\pi_\theta(y^*\mid\xi) = p$, then:
\begin{align*}
  \Delta z_{\xi,y^*} \propto p(1-p),
  \qquad
  \Delta z_{\xi,\bar{y}} \propto -(1-p)p.
\end{align*}

By the chain rule, the full parameter gradient satisfies:
\begin{align*}
\begin{aligned}
\nabla_\theta \mathcal{J}(\theta;\xi)
&\propto 
\frac{\partial z_{\xi,y^*}}{\partial\theta}\,p(1-p)
+
\frac{\partial z_{\xi,\bar{y}}}{\partial\theta}\,\bigl[-p(1-p)\bigr] \\
&= p(1-p)
\left(
  \frac{\partial z_{\xi,y^*}}{\partial\theta}
  -
  \frac{\partial z_{\xi,\bar{y}}}{\partial\theta}
\right).
\end{aligned}
\end{align*}

Define
\begin{align*}
    \mathbf{d}(\xi,\theta)
    :=
    \frac{\partial z_{\xi,y^*}}{\partial\theta}
    -
    \frac{\partial z_{\xi,\bar{y}}}{\partial\theta}.
\end{align*}
We therefore have:
\begin{align*}
    \nabla_\theta\mathcal{J}(\theta;\xi)
    = p(1-p)\,\mathbf{d}(\xi,\theta).
\end{align*}
Taking norms yields:
\begin{align*}
    \|\nabla_\theta\mathcal{J}(\theta;\xi)\|
    = p(1-p)\,\|\mathbf{d}(\xi,\theta)\|.
\end{align*}
For fixed state $\xi$ and parameters $\theta$, the magnitude of the policy gradient is directly proportional to $p(1-p)$, with proportionality constant $\|\mathbf{d}(\xi,\theta)\| > 0$, which completes the proof.
\end{proof}

\section{Significance of “Warmup Training” and “Data Learnability”}
\label{longer_ablation_study}
To fully show the role of "warmup training" and "data learnability", we conducted the ablation experiments on Qwen2.5-3B and Qwen2.5-7B  by training for 2,000 and 1,000 steps, respectively. As shown in Table~\ref{significance_2}, both of them have a significant impact on the proposed method through sufficient training.

\begin{table*}[h!]
\centering
\caption{\label{significance_2}Ablation study of warmup training and data learnability. We train Qwen2.5-3B and Qwen2.5-7B on the DAPO-MATH-17K selected 1,000 examples for 2,000 and 1,000 steps, respectively.} 
\renewcommand{\arraystretch}{0.8}
\setlength{\tabcolsep}{6pt}
\begin{tabular}{@{}c|ccc@{}}
\midrule
\textbf{Benchmark} & \textbf{GSM8K} & \textbf{MATH500} & \textbf{AMC2023} \\
\midrule
\method (Qwen2.5-3B, 2,000 steps) &  83.8 &   67.8  &   36.9\\
 \midrule
w/o warmup training  (Qwen2.5-3B, 2,000 steps)  & 81.9&  64.4 &  31.0\\
w/o data learnability (Qwen2.5-3B, 2,000 steps) & 81.3  & 63.6 & 34.8\\
\midrule
\midrule
 \method (Qwen2.5-7B, 1,000 steps) &  {90.4} & {76.7} &  {48.6}\\
 \midrule
w/o warmup training  (Qwen2.5-7B, 1,000 steps)  & 89.9 &  73.2&  43.7\\
w/o data learnability (Qwen2.5-7B, 1,000 steps) & 89.9 & 75.3 & 46.4\\
\bottomrule
\end{tabular}
\end{table*}




\section{Detailed Discussion on Computational Efficiency}
\label{Detailed Discussion on time}
\addcontentsline{toc}{section}{Appendix: Detailed Discussion on Practical Data Selection Time}
This appendix provides a detailed discussion on the time cost and computational efficiency of LearnAlign. 
We elaborate on (1) complexity analysis, (2) efficient implementation of gradient-information estimation (Step~3), 
(3) efficiency of \textit{LearnAlign} score computation despite the nominal $n \times n$ matrix size (Step~4), 
and (4) a comparison of time costs against baseline methods.
\subsection{Complexity Analysis}
\label{cost_time}
Let \(n = |\mathcal{D}_{\text{train}}|\), \(m = |\mathcal{D}_{\text{warmup}}| \ll n\), and \(d\) be the projected gradient dimension. Let \(C_{\nabla \mathcal{J}}\) and \(C_{\text{gen}}\) denote the time cost of computing one gradient and generating one rollout, respectively.  
The data selection includes four steps:  
\textbf{(1)} RLVR fine-tuning on \(\mathcal{D}_{\text{warmup}}\) to obtain \(\model_s\): time \(\mathcal{O}(m C_{\nabla \mathcal{J}})\), space \(\mathcal{O}(\dim(\model))\).
\textbf{(2)} Generating \(G\) rollouts per sample and computing Learnability: time \(\mathcal{O}(n G C_{\text{gen}})\), space \(\mathcal{O}(n)\). 
\textbf{(3)} Computing GRPO gradients for each \(\sample \in \mathcal{D}_{\text{train}}\) and projecting to \(\phi(\model_s;\sample)\in\mathbb{R}^d\): time \(\mathcal{O}(n C_{\nabla \mathcal{J}})\), space \(\mathcal{O}(nd)\).  
\textbf{(4)} Constructing the pairwise score matrix \(\mathbf{S}\in\mathbb{R}^{n\times n}\) and averaging rows to select top-\(N\): time \(\mathcal{O}(n^2 d)\), space \(\mathcal{O}(n^2)\).

\subsection{Efficient Implementation of Gradient Information Estimation (Step 3)}
\addcontentsline{toc}{subsection}{A2. Efficient Implementation of Gradient Information Estimation}

\paragraph{Current efficiency measures in our method.}

As shown in Table~\ref{tab:time-step}, the Gradient Information Estimation step (Step 3) is the most time-consuming part of our method. We adopt two strategies to make gradient-information estimation efficient:
\vspace{-5pt}
\begin{itemize}[leftmargin=10pt]
    \item \textbf{Single-rollout gradient computation.}
Following~\citep{lin2025cppo}, we compute the gradient of a single correct rollout per sample, which significantly reduces backpropagation cost.  
Table~\ref{time} of the main paper shows that this yields substantial savings while preserving the informative gradient directions required for LearnAlign.

\item \textbf{Random projection of gradients.}
Full-dimensional gradients are prohibitively large.  
We adopt a Johnson--Lindenstrauss–style~\citep{johnson1984extensions} random projection:
\[
\phi(\theta; x) \;=\; \Gamma^\top \nabla \mathcal{J}_{\mathrm{GRPO}}(\theta;x),
\]
\[\text{where}\quad\Gamma \in \mathbb{R}^{k \times d}, \; d \ll k.\]
This preserves inner products, enabling efficient computation of \textit{LearnAlign} scores in a low-dimensional space.
\end{itemize}

\paragraph{Other possible techniques for efficient computation.}
\vspace{-5pt}
\begin{itemize}[leftmargin=10pt]
    \item \textbf{Cancellation effect.} Prior work~\citep{yeh2022first} shows that token-level gradients can exhibit cancellation across time steps, allowing partial reuse of intermediate results and reducing redundant backpropagation.
\item \textbf{LoRA-space gradients.} Instead of backpropagating through the full parameter space, one may compute gradients only within a low-rank LoRA subspace~\citep{hu2022lora}, dramatically reducing dimensionality while preserving informative update directions.
\item \textbf{Neural-network surrogate models for influence prediction.} A potential direction is to train a compact neural network to predict influence scores from cheaper metadata (e.g., embeddings, rollout statistics). Prior studies ~\citep{agarwalneural} show such surrogate models can remove the need to compute full gradients for every sample.
\end{itemize}

\subsection{Efficient Implementation of \textit{LearnAlign} Score Matrix (Step 4)}
\addcontentsline{toc}{subsection}{A3. Efficient Implementation of \textit{LearnAlign} Score Matrix}

\begin{table}[t]
\centering
\caption{Time cost of different steps in LearnAlign.}
\label{tab:time-step}
\resizebox{0.85\linewidth}{!}{
\begin{tabular}{lc}
\toprule
\textbf{Step} & \textbf{Time} \\
\midrule
Step 1: Warmup Training & 2h  2min \\
Step 2: Learnability Computation & 2h 41min \\
Step 3: Gradient Information Estimation & 4h 12min \\
Step 4: LearnAlign-based Data Selection & $<$1 min (12.7s) \\
\midrule
\textbf{Total} & {8h 55min} \\
\bottomrule
\end{tabular}}
\end{table}


Although Step~4 conceptually involves an $n \times n$ \textit{LearnAlign} score matrix, the computation is extremely efficient.
\paragraph{Current data scales ($n=10^3$\textasciitilde$10^4$).}
In our experiments, the training set size is at most a few tens of thousands.  
Step~4 is implemented as a single batched GPU matrix multiplication on low-dimensional gradient features.  
Table~\ref{tab:time-step} shows that \textit{LearnAlign} selection takes only \textbf{12.7 seconds}, compared with over \textbf{4 hours} for gradient computation. The computational bottleneck lies overwhelmingly in obtaining gradients, not in matrix operations.
\paragraph{Scalable extensions for ultra-large datasets.}
When $n$ reaches hundreds of thousands, the following scalable methods can be further applied:
\vspace{-5pt}
\begin{itemize}[leftmargin=10pt]
    \item \textbf{Low-rank/Nyström sampling.} Approximate the full similarity matrix using a small subset of rows/columns (e.g., via Nyström sampling \citep{williams2000using}), reducing cost from $O(n^2)$ to $O(nc)$, where $c$ is the number of sampled rows/columns and $c \ll n$. 
    \item \textbf{Two-stage cascade selection\citep{gong2025two}}: Use a cheap embedding-based filter to reduce the candidate pool, then apply \textit{LearnAlign} only on that smaller set.
\end{itemize}

\subsection{Computational Cost of Baseline Methods}
\addcontentsline{toc}{subsection}{A4. Computational Cost of Baseline Methods}

Table~\ref{tab:baseline-cost} summarizes the time cost of different data-selection baselines for training Qwen2.5-3B on DAPO-MATH-17K.  
All baselines include the same warmup training time (2h2min) and rollout sampling (2h41min). 

\begin{table}[t]
\centering
\caption{Comparison of data selection time cost for different methods.}
\label{tab:baseline-cost}\resizebox{0.5\linewidth}{!}{
\begin{tabular}{lc}
\toprule
\textbf{Method} & \textbf{Time} \\
\midrule
PPL-Top & 5h 34min \\
PPL-Middle & 5h 34min \\
IFD & 6h 26min \\
Token Length & 4h 44min \\
SelectIT & 6h 54min \\
LIMR & 43h 12min \\
\method & {8h 55min} \\
\bottomrule
\end{tabular}}
\end{table}


\section{Discussion on Representativeness and Diversity}
To assess the trade-off between diversity and representativeness, we conduct additional experiments that incorporate feature-space diversity~\citep{xia2024rethinking}. For example, we combine K-means clustering with \textit{LearnAlign}, selecting the highest-scoring samples within each cluster to promote diversity. As shown in Table~\ref{tab:rep_div}, incorporating explicit feature-space diversity does not yield significant gains over \textit{LearnAlign}, which prioritizes representativeness. Moreover, the diversity-aware variant remains sensitive to the choice of the number of clusters $k$.

As reported in LIMO~\citep{ye2025limo}, the reasoning capability stimulated by an example is not directly correlated with shallow features, making traditional diversity criteria (e.g., k-means over embeddings) unreliable.
For RLVR reasoning, recent studies~\citep{ye2025limo,li2025limr} show that very small subsets of high-value reasoning data, even a one-shot example, can provide broad generalization improvements across categories. Overall, current evidence suggests that representative and learnable samples are the primary bottleneck for policy improvement, and feature-level diversity provides limited additional benefit. Therefore, our method prioritizes representativeness. Nevertheless, we acknowledge that diversity-aware RLVR data selection remains underexplored, and investigating principled diversity metrics beyond surface features is an important direction for future work.

\begin{table}[ht]
\centering
\caption{The performance of \textit{LearnAlign} that integrates the K-means clustering on DAPO-MATH-14K with Qwen2.5-3B.}
\label{tab:rep_div}\resizebox{\linewidth}{!}{
\begin{tabular}{lccc}
\toprule
\textbf{Model} & \textbf{GSM8K}&\textbf{MATH500}&\textbf{AMC2023} \\
\midrule
\method&79.3&60.2&{28.3}\\
+k-means ($k$=5)&77.5&59.8&27.1 \\
+k-means ($k$=10)&{80.3}&{60.8}&27.4\\
+k-means ($k$=20)&78.4&60.4&26.8\\
\bottomrule
\end{tabular}
}
\end{table}

\section{Discussion on Influence of Selected Data Size}

To study the influence of selected data size, we select more examples for the GSM8K dataset with Qwen2.5-1.5B-Instruct using \textit{LearnAlign}. As shown in Table~\ref{data size}, a better result can be obtained by selecting a subset with moderate data size.

\begin{table}[t]
\centering
\caption{Comparison of the influence of selected data size for the GSM8K dataset with Qwen2.5-1.5B-Instruct using \textit{LearnAlign}.}
\label{data size}\resizebox{0.55\linewidth}{!}{
\begin{tabular}{c|c}
\toprule
\textbf{Data Size} & \textbf{Test Performance} \\
\midrule
100  & 74.8 \\
500 & 76.4 \\
1,000  & 77.5 \\
2,000  & 78.3 \\
3,000  & 77.9\\
4,000  & 78.7 \\
5,000  & 78.1 \\
6,000   & 77.1 \\
FULL(7,473)  & 77.0 \\
\bottomrule
\end{tabular}}
\end{table}

\section{Sensitivity of Warmup Dataset}
The warmup dataset also may affect the performance of \textit{LearnAlign}. We perform experiments with three different warmup datasets on DAPO-MATH-14K with Qwen2.5-3B. As shown in Table~\ref{tab:sen_warmup_data}, the proposed data selection method is robust to the randomness of the initial warmup dataset.
\begin{table}[ht]
\centering
\caption{The performance of \textit{LearnAlign} with three different warmup datasets on DAPO-MATH-14K with Qwen2.5-3B.}
\label{tab:sen_warmup_data}\resizebox{\linewidth}{!}{
\begin{tabular}{lccc}
\toprule
\textbf{Model} & \textbf{GSM8K}&\textbf{MATH500}&\textbf{AMC2023} \\
\midrule
\method  (warmup dataset 1)&79.3&60.2&28.3\\
\method  (warmup dataset 2)&79.5&61.8&29.5\\
\method  (warmup dataset 3)&81.2&60.4&28.9\\
\bottomrule
\end{tabular}}
\end{table}


\section{Comparison with Filtering Data by  Pass@N Score}

\begin{table*}[ht]
\centering
\caption{Comparison of data selection methods on three benchmarks. We train Qwen2.5-3B on the DAPO-MATH-17K with a selected subset.}
\label{tab:other_baselines}
\begin{tabular}{lccc}
\toprule
\textbf{Data Selection Method} & \textbf{GSM8K}&\textbf{MATH500}&\textbf{AMC2023} \\
\midrule
Qwen2.5-3B-FULL (2,174 steps)&83.6&65.8&31.0\\
\midrule
Learnability (2,000 steps)&82.9&65.0&33.4\\
Pass@8 Score Filter (2,000 steps)&83.5&64.6&31.7\\
\method  (2,000 steps)&{83.8}&{67.8}&{36.9} \\
\midrule
\midrule
Qwen2.5-7B-FULL (1,000 steps)& 90.0 & {77.6} &47.3\\
\midrule
Learnability (1,000 steps)&89.9&74.4&46.4\\
Pass@8 Score Filter (1,000 steps)&89.9&75.0&43.7\\
\method  (1,000 steps)&{90.4}&76.7&{48.6} \\
\bottomrule
\end{tabular}
\end{table*}

In addition to the existing baselines, we also consider a baseline that selects data using the pass@N score, which measures how often a model successfully solves a problem across N independent attempts. Specifically, we implement a pass@8–based filtering strategy: we remove questions whose pass@8 score falls in $\{0,1,7,8\}$, as these correspond to samples that are either extremely easy or extremely difficult. Since the number of samples selected by pass@8–based filtering is not fixed, we train all Qwen2.5-3B and Qwen2.5-7B for about 2,000 steps and 1,000 steps respectively, to ensure a fair comparison.
For completeness, we also include a learnability-only baseline that selects the 1,000 samples with the highest success-rate-based learnability introduced in this paper.

The comparison is shown in Table~\ref{tab:other_baselines}. Furthermore, we also highlight two key observations based on the actual results:
\begin{enumerate}[leftmargin=10pt]
    \item  Learnability and pass@8 filtering achieve comparable performance by selecting medium-difficulty samples. Both methods aim to avoid overly easy and overly hard questions, and therefore select samples near the “middle” of the model’s current capability. This results in comparable performance between the two methods across three benchmarks. Importantly, both methods achieve accuracy relatively close to full-data training, confirming the intuition that medium-difficulty samples carry substantial training value under RLVR.
    \item \textit{LearnAlign} outperforms full-data training and clearly surpasses baselines. In contrast to purely difficulty-based filtering, \textit{LearnAlign} incorporates gradient-direction alignment to additionally capture the representativeness of each sample. As shown in Table~\ref{tab:other_baselines}, \textit{LearnAlign} exceeds the full-data baseline and shows a clear margin over both Learnability and Pass@8 filtering across all three benchmarks. This demonstrates that combining learnability with gradient alignment signals yields a substantially more informative subset than using difficulty signal alone.
\end{enumerate}

Overall, the results indicate that while selecting medium-difficulty samples is beneficial, considering gradient alignment is essential for identifying the truly most impactful RLVR data, leading to stronger and more consistent gains. Besides, we are running additional experiments and will update more results once we finish them.

\color{black}
\section{Staged Reinforcement Learning with LearnAlign}
\begin{figure}[h]
		\centering
\includegraphics[width=0.95\linewidth]{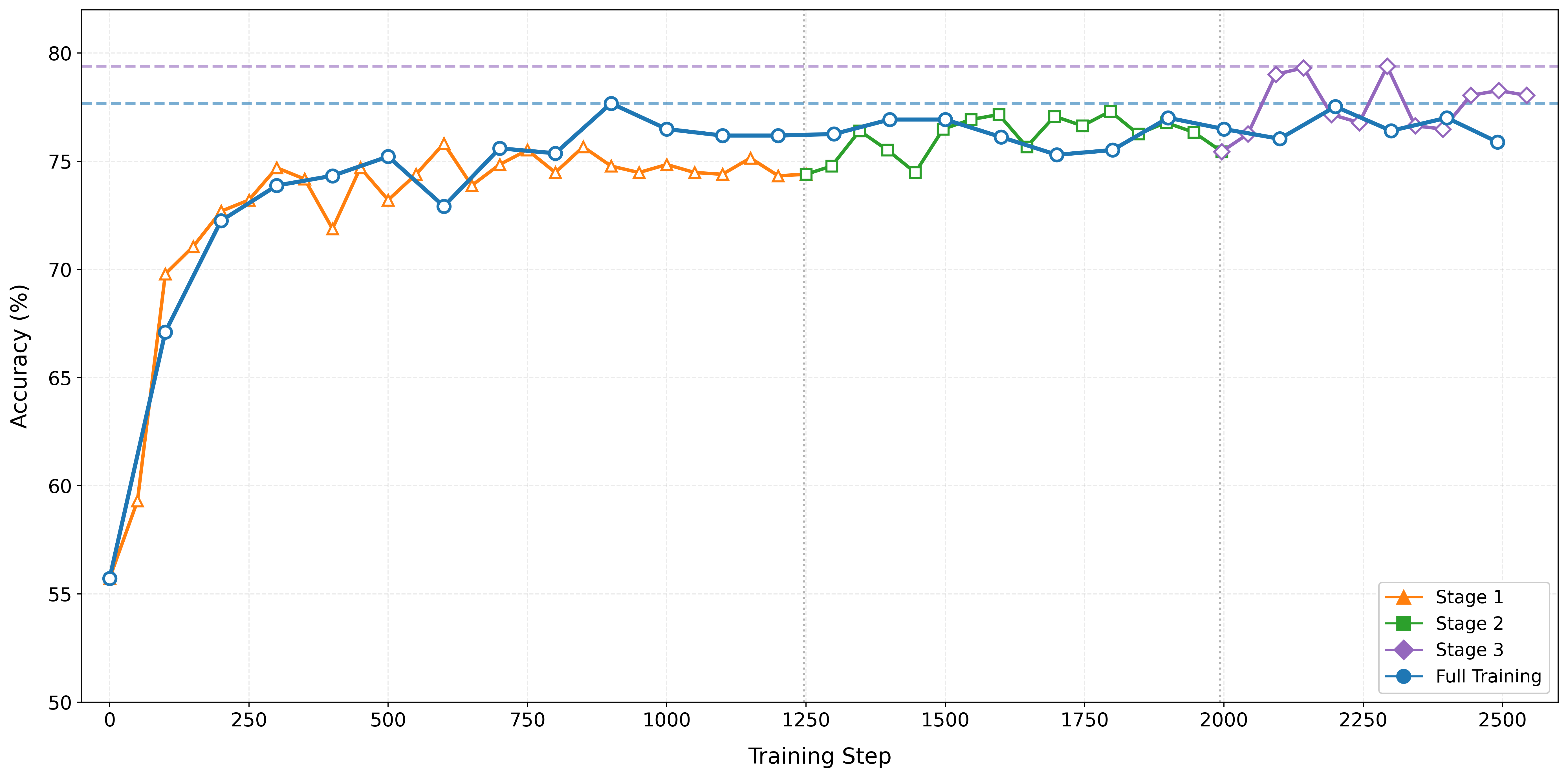}
		\caption{The performance of the staged reinforcement learning with the proposed data selection method.}
		\label{fig2}
	\end{figure}
\label{staged}
To further assess the applicability of our method in a curriculum learning scenario, we design a three-stage training procedure on the GSM8K dataset. Specifically:
\begin{itemize}
  \item In the \textbf{first stage}, we use {Qwen2.5-1.5B-Instruct} to select the top {50\%} of the training samples, and train the model until convergence.
  \item In the \textbf{second stage}, the resulting model is used to select the next {30\%} of the samples, and again trained to convergence.
  \item In the \textbf{final stage}, the latest model selects the  {20\%} of samples, and training is repeated until convergence.
\end{itemize}

As shown in Figure~\ref{fig2}, our method can seamlessly integrate into such a staged RLVR curriculum to significantly improve capability acquisition. 

\section{Case Study}
In this section, we provide case studies on the GSM8K benchmark with Qwen2.5-1.5B-Instruct. As shown in the following cases, the Qwen2.5-1.5B-Instruct model trained on the full dataset has illogical reasoning, leading to an incorrect answer, while our method derived the correct answer through clear and logical steps.
\begin{exbox}{Qwen2.5-1.5B-Instruct-FULL:}
\textbf{[Questions:]}
Sophia is thinking of taking a road trip in her car, and would like to know how far she can drive on a single tank of gas. She has traveled 100 miles since last filling her tank, and she needed to put in 4 gallons of gas to fill it up again. The owner's manual for her car says that her tank holds 12 gallons of gas. How many miles can Sophia drive on a single tank of gas?

\textbf{[Answer:]}
<think> Let's break it down step-by-step. Sophia has traveled 100 miles and took 4 gallons of gas to fill up the tanks. The total tank capacity is 12 gallons. First, we need to find if Sophia has reached the 12-gallon mark yet after traveling 100 miles. We do that by dividing the total distance by the tank's capacity: 100 miles ÷ 12 gallons = approximately 8.33 gallons used to travel the first 100 miles. Since it took her 4 gallons to refill, it means Sophia used around 4 gallons for the initial 100 miles, so her current tank usage allows her to travel 4 gallons divided by 4 gallons per tank used = 1 tank away from the 12-gallon mark. After that, she'll be able to travel another 12 - 4 = 8 gallons of fuel. </think><answer> Sophia can drive 8 more miles on a single tank of gas. </answer> <think> Therefore, the answer is 8. </think> <answer> 8 </answer>"
\end{exbox}

\begin{exbox}{\textit{LearnAlign} (subset with 2000 examples)}
\textbf{[Questions:]}
Sophia is thinking of taking a road trip in her car, and would like to know how far she can drive on a single tank of gas. She has traveled 100 miles since last filling her tank, and she needed to put in 4 gallons of gas to fill it up again. The owner's manual for her car says that her tank holds 12 gallons of gas. How many miles can Sophia drive on a single tank of gas?

\textbf{[Answer:]}
<think> Sophia has already traveled 100 miles since the last tank was filled, and she needed 4 gallons of gas to refill the tank. To calculate the car's fuel efficiency, we divide the miles traveled by the gallons of gas used. In this case, the fuel efficiency is 100 miles / 4 gallons = 25 miles per gallon. Now, to calculate how far the car can drive on a full tank, we multiply the number of gallons the tank holds by the car's fuel efficiency. In this case, 12 gallons * 25 miles per gallon = 300 miles. </think><answer> 300 </answer>
\end{exbox}

\end{document}

%% file: main/intro.tex
\section{Introduction}

Recently, reinforcement learning with verifiable rewards (RLVR)  has become a successful and crucial post-training paradigm for enhancing the reasoning ability of large language models (LLMs), exemplified by OpenAI o1~\citep{jaech2024openai}, DeepSeek-R1~\citep{guo2025deepseek}, Kimi k1.5~\citep{team2025kimi}, and so on. These models commonly utilize a rule-based reward function, such as the correctness of mathematical solving and code generation problems, to provide supervision signals.

    	\begin{figure*}[ht]
		\centering
\includegraphics[width=0.95\linewidth]{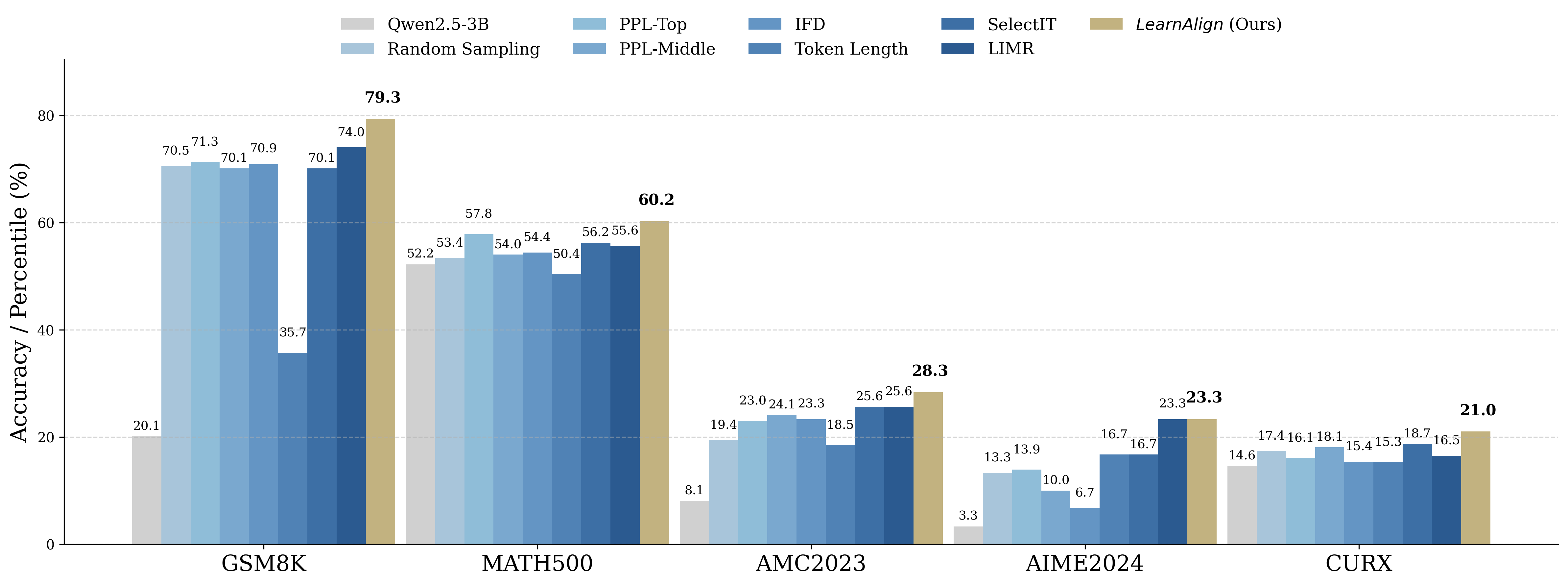}
		\caption{Performance comparison between baseline methods and our proposed \textit{LearnAlign} on various benchmarks, including GSM8K, MATH500, AMC2023, AIME2024, and CRUX, using the Qwen2.5-3B model.}
		\label{fig1}
	\end{figure*}
Due to the large number of parameters, the post-training for LLMs usually needs a lot of computing resources with large-scale data~\citep{zhou2023lima,luo2024mmevol,luo2025gui,liu2025mtp,li2022selective,li2022estimating,zhang2024m3d,zhang2024dance}. While, according to the recent studies~\citep{zhou2023lima,ye2025limo}, it is feasible to 
activate the specialized ability of a pre-trained language model in downstream tasks with a small set of examples. Inspired by this observation, several works~\citep{xia2024less,li2023quantity,liu2024selectit} have explored data selection strategies for the post-training of LLMs. Most of methods assign a quality score to each sample using an external expert model or the target model's training signals, then select the top-ranked data. While, these works are usually designed for the supervised fine-tuning (SFT) paradigm rather than the reinforcement learning paradigm, which shows limited effectiveness in reasoning-oriented scenarios. As far as we know, there are very few works~\citep{li2025limr,wang2025reinforcement} that studied the data selection problem of the reinforcement learning paradigm at present. These works~\citep{li2025limr,wang2025reinforcement} verified that a small amount of data or even one training example can still provide sufficient information for RLVR post-training. However, their methods are not computationally efficient, since they need to train the whole original dataset for several epochs during data selection, which makes them less practical for saving computing resources.
    
To address the above issue, we propose a practical data selection method, named \textit{LearnAlign}, for the RLVR paradigm in large language models via gradient alignment. Inspired by~\citep{pruthi2020estimating,xia2024less}, to select the high-valued reasoning data, we consider measuring the influence of each data point for training the LLM. First, we estimate the influence of one data point for the training dataset by approximating the change of the training loss using a first-order Taylor expansion. Such influence then can be transformed to the alignment score of gradients between that data point and the training dataset, which can reflect the representativeness of data points to the dataset. In addition, to address the  well-known response-length bias for gradient norms~\citep{liu2025understanding,xia2024less},  we introduce the learningability of data estimated by the success rate to replace it, which can represent the learnable potential without the bias~\citep{florensa2018automatic,tzannetos2023proximal}. Finally, we can obtain an improved gradient alignment score, and then the top-sorted data points are identified as the learnable and representative reasoning data. 

Experiments across four mathematical reasoning benchmarks (GSM8K~\citep{cobbe2021training}, MATH500~\citep{hendrycks2021measuring}, AMC2023~\citep{AMC2023mathematical}, and AIME2024~\citep{AIME2024mathematical}) and one code generation benchmark (CRUX~\citep{gu2024cruxeval}) reveal two key findings: (1) conventional SFT data selection methods fall short in the RLVR paradigm for the post-training phase of LLMs; (2) \textit{LearnAlign} achieves minor performance degradation or even superior performance while requiring only a fraction of the training data (as seen in Figure~\ref{fig1}). Notably, our method achieves comparable performance compared to full data (42.4\% vs. 44.9\%) using much less data (1,000 vs. 17,000 examples) across five benchmarks.
Our main contributions are summarized as follows:
\vspace{-5pt}
\begin{itemize}[leftmargin=10pt]
\item In this paper, we explore efficient data selection for RLVR post-training from the perspective of gradient alignment, a direction that has received limited attention in prior work.
\item We introduce \textit{LearnAlign}, a novel data selection framework that constructs learnability-weighted gradient representations to measure influence between data points, where the learnability metric captures learning potential and addresses the response-length bias for gradient norms.
\item Comprehensive comparison with prior methods across five benchmarks and three LLMs clearly reveals the shortcomings of traditional SFT data selection methods, and demonstrates that LearnAlign identifies high-value subsets that match or exceed full-dataset performance.
\end{itemize}

%% file: main/related.tex
\section{Related Work}
We review the existing data selection studies for LLM post-training, including supervised fine-tuning (SFT) and reinforcement learning with verifiable rewards (RLVR).

\paragraph{\textbf{Data selection for SFT post-training:}}
Most data selection methods for LLM supervised fine-tuning assign quality scores to each sample using various signals, which we categorize into external-scoring and self-scoring approaches.
External-scoring methods often rely on powerful LLMs. For example, INSTAG~\citep{lu2023instag} uses ChatGPT to generate fine-grained tags for assessing instruction diversity and complexity; ALPAGASUS~\citep{chen2023alpagasus} employs ChatGPT to directly evaluate and select high-quality instructions; IFD~\citep{li2023quantity} measures discrepancies between model and self-generated outputs; and LESS~\citep{xia2024less} prioritizes data via a gradient-based similarity to few-shot examples.
Self-scoring methods avoid external models. SelectIT~\citep{liu2024selectit} uses multi-level uncertainty (token, sentence, and model) to identify high-quality data. Nuggets~\citep{li2023one} scores samples by their influence on perplexity of an anchor set, improving tuning efficiency.

\paragraph{\textbf{Data selection for RLVR post-training:}} As far as we know, there are few works that have explored data selection for RLVR post-training. LIMR \citep{li2025limr} and 1-shot RLVR~\citep{wang2025reinforcement} verify earlier that a small amount of data can still provide sufficient information for the RLVR training. While these methods are not computationally efficient, since they need to train the original dataset for several epochs during data selection. To address this issue, this work offers a more practical solution for RLVR post-training.

%% file: main/preliminary.tex
\section{Preliminary}

A next-token prediction LLM can be regarded as a token-level Markov Decision Process (MDP) \citep{sutton1998reinforcement,foster2025learning}, which is denoted by a tuple \(\mathcal{M}:=\left\{\mathcal{S},\mathcal{A}, \gamma, \mathcal{T}, \mathcal{R}, \mathcal{P}^0\right\}\).  \(\mathcal{S}\) represents the state space, and \(\mathcal{A}\) denotes the action space. \(\mathcal{P}^0\) means the starting state distribution while \(\mathcal{T}\) is the transition function.  The reward function and the discount factor are denoted \(\mathcal{R}\) and \(\gamma\), respectively. LLM post-training by RLVR is formulated as a token-level MDP, where the objective is to sequentially generate text conditioned on the given prompt. It starts from a prompt or question query denoted as \(\sample=[\xi_1, \xi_2,\cdots,\xi_n]\), represents \(n\) tokens. At each timestep \(t\), the action \(y_t \in \mathcal{A}\) corresponds to the generation of a token \(y_t\), sampled from the model's output distribution. The transition function \(\mathcal{T}([\sample_{0:t-1}, y_t]) = \sample_{0:t}\) is deterministic. It concatenates the generated token \(y_t\) to the existing sequence \(\sample_{0:t-1} = [\xi_1, \dots, \xi_n, y_1, \dots, y_{t-1}]\) to form the new state \(\sample_{0:t} = [\xi_1, \dots, \xi_n, y_1, \dots, y_t]\).  The reward for generating token \(y_t\) at timestep \(t\) is sparse, assigned only at the final timestep \(T\) of the episode. The reward is binary, with \(\mathcal{R}(\sample\mathbf{y}) = 1\) if the complete sequence \(\sample\mathbf{y} = [\xi_1, \dots, \xi_n, y_1, \dots, y_T]\) (the prompt followed by the generated tokens) is correct, and \(\mathcal{R}(\sample\mathbf{y}) = 0\) otherwise. Typically, the discount factor \(\gamma\) is set to 1 , so the cumulative discounted finite-horizon return is simply \(\mathcal{R}(\sample\mathbf{y})\).

\paragraph{\textbf{Group relative policy optimization (GRPO).}} Recently, GRPO  \citep{shao2024deepseekmath} emerges as a popular RLVR algorithm. In particular, the GRPO consists of two terms, a policy term \(\mathcal{J}_{\text{Policy}}\) and another KL divergence term. This can be formulated as follows:


\begin{align}
\begin{aligned}
&\mathcal{J}_{\text{GRPO}}(\theta) = 
\mathbb{E}_{q\sim\mathcal{P}_q,\ \{o_i\}_{i=1}^{G}\sim\pi_{\theta_{\text{old}}}(o\mid q)}
\biggl[
\frac{1}{G} \sum_{i=1}^{G}\\& \frac{1}{|o_i|} \sum_{t=1}^{|o_i|}
\min \Bigl( r_{i,t} \hat{A}_{i,t},\delta\hat{A}_{i,t} \Bigr)- \beta \mathbb{D}_{\text{KL}}[\pi_{\theta} \mid \pi_{\text{ref}}]
\biggr],
\end{aligned}
\label{eq:GRPO}
\end{align}
where $r_{i,t}= \frac{\pi_{\model}(o_{i,t}|q,o_{i,<t})}{\pi_{\model_{\text{old}}}(o_{i,t}|q,o_{i,<t})}$, and \(\hat{A}_{i,t}\) denotes the relative advantage, which is computed using a group of rewards \(\{r_1, r_2,\cdots,r_G\}\):
    $\hat{A}_{i,t} = \frac{r_i - \text{mean}(\{r_i\}_{i=1}^{G})}{\text{std}(\{r_i\}_{i=1}^{G})}.$
$\mathbb{D}_{\text{KL}}$ denotes the KL-divergence between \(\pi_{\model}\) and \(\pi_{\text{ref}}\) to constrain the divergence between the old and new policy model, while \(\delta=\text{clip}(r_{i,t}, 1-\varepsilon,1+\varepsilon)\).
\(\pi_{\text{ref}}\) typically represents the original pre-trained model prior to the RLVR post-training process. 

%% file: main/method.tex
\section{Method}

Here, we outline our strategy for selecting data to effectively enhance the large language model's performance during the reinforcement learning (RL) post-training phase. We begin by defining the data selection problem (Section~\ref{sec:pro_def}). Next, we discuss data influence estimation via gradient alignment (Section~\ref{sec:data_influ_est}) and improving gradient alignment with data learnability (Section~\ref{sec:grad_align}), which provides a way to assess the utility of data pairs. Finally, we present a comprehensive overview of our data selection method (Section~\ref{sec:data_selec_overall}).

    

\subsection{Problem Definition}
\label{sec:pro_def}
The objective of data selection for LLM RLVR post-training is to identify a subset \(\mathcal{D}^s_{\text{train}}\) from the full training dataset \(\mathcal{D}_{\text{train}}\), where \(|\mathcal{D}^s_{\text{train}}| < |\mathcal{D}_{\text{train}}|\). The selected subset is used to train an LLM policy model \(\pi_{\model}\) via reinforcement learning techniques, e.g., PPO \citep{schulman2017proximal} or GRPO \citep{shao2024deepseekmath}, aiming to achieve lower loss and improved performance on a test dataset \(\mathcal{D}_{\text{test}}\). Moreover, no additional information beyond the original training dataset \(\mathcal{D}_{\text{train}}\) is available. Ideally, the selected subset should enable the model to achieve performance comparable to training on the full dataset \(\mathcal{D}_{\text{train}}\) with significantly fewer data, or ensure that any performance degradation is minimal, thereby maximizing training efficiency.

\subsection{Data Influential Estimation via Gradient Alignment}
\label{sec:data_influ_est}
Similar to SFT data selection methods~\citep{xia2024less}, selecting data for LLM post-training also requires analyzing and understanding the training dynamics of the data. Specifically, we need to identify which data can most effectively reduce the model's loss. Drawing inspiration from \citep{pruthi2020estimating,liu2024less}, the change in the loss function \(\mathcal{J}(\cdot)\) for a given data \(\sample'\) as the model parameters change from \(\model^t\) to \(\model^{t+1}\) can be approximated using a first-order Taylor expansion as follows:
\begin{align}
\begin{aligned}
    \mathcal{J}(\model^{t+1};\sample')  &\approx\mathcal{J}(\model^t;\sample')\\+&\nabla\mathcal{J}(\model^t, \sample')(\Delta\model) 
    +\mathcal{O}(\|\Delta\model\|^2).
    \end{aligned}
    \label{eq:taylor_expan}
\end{align}
where \(\Delta\model=\model^{t+1}-\model^t\)If the model \(\model^{t+1}\) is trained by a single data \(\sample\) with stochastic gradient descent (SGD) at time \(t\), this can be expressed as \(\model^{t+1} = \model^{t}-\eta_t\nabla\mathcal{J}(\model^t;\sample)\), where \(\eta_t\) denotes the learning rate for the time \(t\). Substituting this update into Eq.(\ref{eq:taylor_expan}), a data \(\sample\) update to the model introduces the change of the loss on another sample \(\sample'\), which can be formulated as:
\begin{align}
\begin{aligned}
    \mathcal{J}&(\model^{t+1};\sample')- \mathcal{J}(\model^t;\sample')
    \approx\nabla\mathcal{J}(\model^t;\sample')\Delta\model\\
    &=-\eta_t\left(\nabla\mathcal{J}(\model^t;\sample') \cdot \nabla\mathcal{J}(\model^t;\sample)\right),
\end{aligned}
\end{align}
where we ignore the higher-order term \(\mathcal{O}(\|\model^{t+1}-\model^t\|^2)\) as it is small for a sufficiently small step size $\eta_t$. Based on this, we can formalize the influence between two data \(\sample_i\) and \(\sample_j\).
\begin{definition}[Data Influence via Gradient Alignment]
Let \(\sample_i\) and \(\sample_j\) be two data from the training dataset \(\mathcal{D}_{\text{train}}\), and let \(\model\) represent the model parameters. The influence of data \(\sample_i\) on data \(\sample_j\), denoted as \(\text{Inf}_t(\sample_i, \sample_j)\), is defined as the dot product of the gradients of the loss function \(\mathcal{J}(\cdot)\) with respect to the model parameters, evaluated at \(\model^t\):
\begin{align}
    \text{Inf}_t(\sample_i, \sample_j) = \nabla \mathcal{J}(\model^t; \sample_i) \cdot \nabla \mathcal{J}(\model^t; \sample_j).
\end{align}
This quantity measures the first-order effect of updating the model with data \(\sample_i\) on the loss of data \(\sample_j\), capturing the similarity in their training dynamics.
\end{definition}
The gradients for each data point reflect the average gradients of all tokens within that data. Previous studies have observed that the gradient norm is inversely correlated with response length~\citep{liu2025understanding,xia2024less}. Using only the inner product of gradients between two data points may bias the data selector toward shorter sequences. To address this issue, some works \citep{wang2020optimizing,xia2024less} employ the cosine similarity instead, but they still suffer from performance degradation when selecting data for post-training LLMs.

\begin{figure*}
    \centering
    \includegraphics[width=1\linewidth]{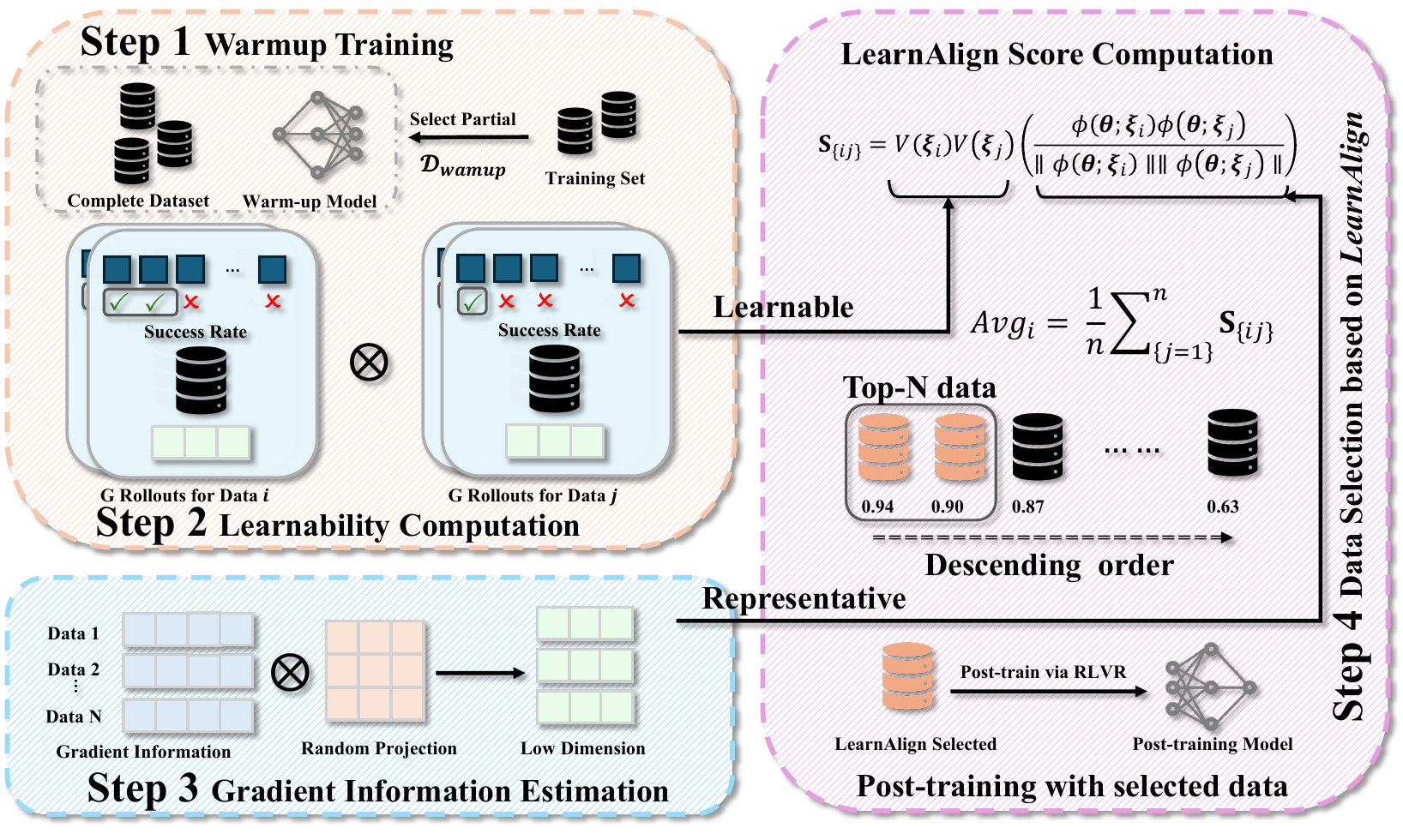}
    \caption{The procedure of the proposed selection method based on improved gradient alignment. We obtain the gradient information and learnability of each data point through Steps 1-3 and then select data for subsequent training according to datapoint-wise \textit{LearnAlign} score in Step 4.}
    \label{fig:enter-label}
\end{figure*}

\subsection{Improving Gradient Alignment with Learning Potential}
\label{sec:grad_align}
\textbf{Motivation:} Based on the preceding analysis, the post-training dynamics of large language models reveal two critical limitations when using the cosine similarity of data gradients as a selection criterion:
(1) \textbf{Loss of Magnitude Information.} By normalizing the gradients, the cosine similarity focuses exclusively on their directional alignment, thereby discarding magnitude information. In post-training LLMs, the gradient magnitude often indicates a data point's influence on model updates, which is essential for effective policy optimization. Ignoring this aspect prevents the cosine similarity from prioritizing data that could drive more substantial improvements in model performance.
(2) \textbf{Failure to Capture Learning Potential.} The cosine similarity does not account for the learning potential of data. Even if two data points exhibit high directional similarity, their utility may be limited if they are either too easy (success rate \( p \approx 1 \)) or too difficult (success rate \( p \approx 0 \)) for the current policy, leading to suboptimal data selection. This limitation aligns with the theory of the Zone of Proximal Development (ZPD) \citep{chaiklin2003zone}, which suggests that effective learning occurs when tasks are of moderate difficulty—neither too challenging nor too simple—for the learner (e.g., an LLM).

To address the aforementioned limitations, we introduce a data learnability metric based on the success rate \(p\), drawing inspiration from prior work to account for both the learning potential and the magnitude of the data \citep{florensa2018automatic, tzannetos2023proximal,foster2025learning}.
\begin{definition}[Data Learnability]
\label{def:learnability}
Consider a sample \( \sample \) evaluated by an LLM policy \( \pi_{\model} \). Let \( p \in [0, 1] \) represent the success rate, defined as the fraction of successful outcomes for the query \( \sample \) across \( G \) rollouts, where \( p \) reflects the probability of a successful learning outcome. The data learnability of data \( \sample \), denoted \( V(\sample) \), is defined as:
\[
V(\sample) = p(1 - p),
\]
where \( 1 - p \) represents the potential for improvement, and \( p(1 - p) \) quantifies the expected learnability of data. This measure captures the sample's utility for enhancing the policy \( \pi_{\model} \), reaching its maximum when \(p=0.5 \), indicating a sample at the boundary of the policy's current capability. Besides, the theoretical justification for the data learnability can be found in Appendix~\ref{Justification} and~\ref{theorem} .
\end{definition}

Built upon the above motivation and our definition of data learnability, we first define a new learnability-weighted gradient vector for each data point \(\sample_i\) as:
\begin{align}
\mathbf{V}(\sample_i) = \frac{\nabla \mathcal{J}(\model; \sample_i)}{\|\nabla \mathcal{J}(\model; \sample_i)\|} \cdot V(\sample_i),
\end{align}
where the first term is the unit gradient vector and \(V(\sample_i)\) is the learnability score (Definition~\ref{def:learnability}). Using these vectors, we can then compute the \textit{LearnAlign} Score between two data points \(\sample_i\) and \(\sample_j\) as
\begin{align}
\begin{aligned}
&\textit{LearnAlign}(\sample_i, \sample_j) = \mathbf{V}(\sample_i) \cdot \mathbf{V}(\sample_j)\\
&= V(\sample_i) V(\sample_j) \frac{\nabla \mathcal{J}(\model; \sample_i)^\top \nabla \mathcal{J}(\model; \sample_j)}{\|\nabla \mathcal{J}(\model; \sample_i) \| \|\nabla \mathcal{J}(\model; \sample_j)\|}.
\label{eq:data_sim}
\end{aligned}
\end{align}
This formulation leverages the learnability of each data point to  weight the gradient inner product by the learning potential, thus reducing the tendency to favor shorter sequences. 


\subsection{Data Selection for RLVR Post-training}
\label{sec:data_selec_overall}
As shown in Figure~\ref{fig:enter-label}, the procedure to select suitable data for RLVR consists of four steps, where we elaborate \textit{LearnAlign} from step 1 to step 4 in detail.

\paragraph{\textbf{Step 1. Warmup Training:}} Initially, we randomly select a small subset \(\mathcal{D}_{\text{warmup}} \subset \mathcal{D}_{\text{train}}\) from the training dataset to perform warmup training on the policy model \(\pi_{\theta}\). This step ensures a more stable and accurate gradient estimation, resulting in a warmed-up model \(\model_s\).

\paragraph{\textbf{Step 2. Learnability Computation:} } We first sample $G$ rollouts for each question and compute the success rate of question $i$ based on the ground truth answer $\mathbf{y}^*$ and the generated answers $\mathbf{y}$ across these $G$ rollouts. The success rate $p$ is calculated as $p=\frac{1}{G}\sum_{g=1}^{G}\mathbb{I}(\mathbf{y}_g=\mathbf{y}^*)$, where $\mathbb{I}$ is the indicator function. Following Definition~\ref{def:learnability}, we can get the learnability $V(\xi_i)$ for each data $i$.

\paragraph{\textbf{Step 3. Gradient Information Estimation:}} Additionally, we can derive the original gradient information from the model \(\model\) checkpoint during the warmup phase of RLVR-based LLM post-training (e.g., GRPO) as follows:
\begin{align}
\begin{aligned}
        &\nabla_{\model}\mathcal{J}_{\text{GRPO}}(\model)= \mathbb{E}_{q\sim\mathcal{P}_{q},\{o_i\}^{G}_{i=1}\sim\pi_{\model_{old}}(o|q)}\Big\{\frac{1}{G} \\
        &  \sum_{i=1}^{G}\frac{1}{|o_i|} \sum_{t=1}^{|o_i|}G(q,o,t,\pi_{\model})\nabla_{\model}\log\pi_{\model}(o_{i,t}|q,o_{i,<t})\Big\},
\end{aligned}
\end{align}
where \(G(q,o,t,\pi_{\model})\) denotes the gradient coefficient \(\hat{A}_{i,t}+\beta\left(\frac{\pi_{\text{ref}}(o_{i,t}|q,o_{i,<t})}{\pi_{\model}(o_{i,t}|q,o_{i,<t})}-1\right)\),  \(\hat{A}_{i,t}\) is computed as GRPO. Since this gradient has nearly the same dimensions as the original model, it is computationally complex. Following prior work, we apply a random projection \(\Gamma\) to the gradient information for each data point \citep{johnson1984extensions, xia2024less}. So we can get a low-dimensional gradient-related information denoted as \(\phi(\model;\sample)=\Gamma^\top\nabla\mathcal{J}_{\text{GRPO}}(\model;\sample)\). 

\paragraph{\textbf{Step 4. Data Selection based on \textit{LearnAlign}:}} Based on the projected gradient from the warmed-up model \(\model_s\), we can rewrite the \textit{LearnAlign} Score between two data \(\xi_i\) and \(\xi_j\) as:
\begin{equation}
\label{eq:final_data_sim}
    S_{ij} =    V(\sample_i) V(\sample_j) \left(\frac{\phi(\model; \sample_i) \phi(\model; \sample_j)}{\|\phi(\model; \sample_i) \| \|\phi(\model; \sample_j)\|}\right).
\end{equation}
So we can get a \(n \times n \) \textit{LearnAlign} Score Matrix \(\mathbf{S}\) (where \(|\mathcal{D}_{\text{train}}|=n\)), capturing the pairwise relation among all data points in the training dataset. Using the \textit{LearnAlign} Score Matrix \(\mathbf{S}\), we select the top-N data. For each data \(\xi_i\), the average \textit{LearnAlign} Score across its row as \(\text{Avg}_i = \frac{1}{n} \sum_{j=1}^n S_{ij}\), where \(S_{ij}\) represents the pairwise alignment scores for all \(j\) (including \(j = i\)) and \(|D_{\text{train}}| = n\). These average scores are then sorted in descending order, and the top-N samples with the highest averages are selected, ensuring the chosen data exhibit the strongest learnability within the training dataset.

%% file: main/exp.tex
\section{Experiments}
We introduce the experimental setup (Section~\ref{setup}) of \textit{LearnAlign}, and then we present the main results (Section~\ref{Results}) on the five benchmarks with some key observations. Moreover, we give some discussions (Section~\ref{ablation}).

\subsection{Experimental Setup}
\label{setup}

\paragraph{\textbf{Settings:}}

We validate the effectiveness of our algorithm under two primary configurations:
{\textbf{(1)}} We train models on subsets of the {GSM8K}~\citep{cobbe2021training} training set with varying sizes: {100}, {500}, {1,000}, and {2,000} samples. The base policy model is {Qwen2.5-1.5B-Instruct}, and evaluation is performed on the GSM8K test set, with greedy decoding used during the inference stage, and the pass@1 accuracy is reported.
{\textbf{(2)}} We train on {1,000} samples from the DAPO-MATH-17K dataset~\citep{yu2025dapo}
 training set using {Qwen2.5-3B} and {Qwen2.5-7B} as the initial policy model. Evaluation is conducted on both four math reasoning benchmarks (GSM8K~\citep{cobbe2021training}, MATH500~\citep{hendrycks2021measuring}, AMC2023~\citep{AMC2023mathematical}, and AIME2024~\citep{AIME2024mathematical}) and one code generation benchmark (CRUX~\citep{gu2024cruxeval}). For GSM8K, MATH500, and CRUX, we report the {pass@1} accuracy; for AMC2023, we report avg@8 as the metric; for AIME2024, we report the pass@8 accuracy. The evaluation temperature is set to 0.8, and the top-p is set to 0.95.
 


\input{main/tabs/main_res}
\input{main/tabs/main_res2}
\paragraph{\textbf{Implementation details:}}
In these experimental settings, for the training hyperparameters, during exploration, we generated 8 rollouts per sample at a temperature of \(1.0\); the learning rate was set to \(1.0\times10^{-6}\); the KL coefficient \(\beta\) was fixed at \(0.04\); and the clipping parameter \(\epsilon\) was set to \(0.2\).
The batch size is set to 48 for GSM8K and 64 for DAPO-MATH-17K. We follow~\citep{xia2024less} for the projection of gradients and use 300 and 1000 samples for warmup training in GSM8K and DAPO-MATH-17K, respectively. For DAPO-MATH-17K, inspired by~\citep{lin2025cppo}, we calculate the gradient of one correct rollout for each sample. 
Additional details are provided in Appendix~\ref{Additional Experimental Details}.

\paragraph{Baselines:} We compare with several baselines: \textbf{Random Sampling}, \textbf{PPL-Top}~\citep{laurenccon2022bigscience}, \textbf{ PPL-Middle}~\citep{ankner2024perplexed}, \textbf{IFD}~\citep{li2023quantity}, \textbf{Token Length}~\citep{xia2024rethinking}, \textbf{SelectIT}~\citep{liu2024selectit}, and \textbf{LIMR}~\citep{li2025limr}.  For GSM8K, we utilize the official solutions in the training data as responses to calculate the above selection metrics. For DAPO-MATH-17K, we make the warmed-up model to generate one response for each problem to conduct their selection. More details about the baselines can refer to Appendix~\ref{appendix:detailed_baselines}.


\subsection{Main Results}
\label{Results}
Table~\ref{tab:main_results} presents the evaluation results of training models on the GSM8K dataset with varying selected data sizes. Table~\ref{tab:main_results2} shows the evaluation results of training models on the DAPO-MATH-17K dataset. 
From these results, we have the following key observations:

\paragraph{\textbf{Key observation 1: Traditional SFT data selection methods fall short in the RLVR paradigm for the post-training phase of LLMs.}} On the one hand, as shown in Table~\ref{tab:main_results}, when the official solutions of the training data are applied as the responses in data selection, traditional SFT approaches show limited and inconsistent effectiveness when applied to RLVR post-training. 
For example, Token Length performs well at 1,000 samples (76.2\%) but drops at 2,000 samples (75.6$\%$). 
On the other hand, as shown in Table~\ref{tab:main_results2}, when the rollouts of the warmed-up model are generated for data selection, PPL-Top is slightly higher than Random Sampling on average. 
Note that none of these baselines consistently outperforms Random Sampling across five benchmarks. Such suboptimal performance of SFT data selection methods may stem from a misalignment between SFT and RLVR objectives. SFT post-training aims to maximize the likelihood of target outputs, where harder examples identified by those methods are often more valuable (assuming they are not noisy). RLVR post-training optimizes for reward maximization, requiring the difficulty to match the model’s current capability.

\paragraph{\textbf{Key observation 2: \textbf{\textit{LearnAlign}}  achieves minor performance degradation or superior performance while requiring only a fraction of the training data.}} As shown in Table~\ref{tab:main_results}, our approach consistently outperforms baselines at every data scale, achieving comparable or superior performance to the full-data training with a small amount of the data. Specifically, With 1,000 samples ($\approx13.4\%$ of full data), {\textit{LearnAlign}}  reaches 77.5\%, already matching the full-data baseline (77.0\%). With 2,000 samples ($\approx26.8\%$ of full data), the proposed method significantly surpasses the full-data training (78.3\% vs. 77.0\%). Besides, with fewer samples (e.g., 100 and 500), the proposed data selection method can largely improve the base model (55.7\%) and even exceed other baselines with more samples, proving that smart selection is better than brute-force scaling, i.e., RLVR post-training with a carefully curated seed set can rapidly unlock a pretrained model’s reasoning ability~\citep{li2025limr}.

\paragraph{\textbf{Key observation 3: \textit{LearnAlign} shows consistent effectiveness across various settings.}} As shown in Table~\ref{tab:main_results} and Table~\ref{tab:main_results2}, our proposed data selection method demonstrates consistent SOTA performance not only on in-distribution (GSM8K, MATH500) but also on out-of-distribution (AMC2023, AIME2024) test sets, and it even generalizes well on the code domain benchmark (CRUX).
In addition, as shown in Appendix~\ref{staged}, \textit{LearnAlign} boosts the performance of RLVR post-training in the staged setting. These results show that it can be effectively applied in various settings by considering learnability and alignment.


    \begin{figure}[t]
    
	\centering
	\subfloat[Gradient norms of examples negatively correlate with the length of the response.]{\includegraphics[width=2.2in,trim=0 10 0 0,clip]{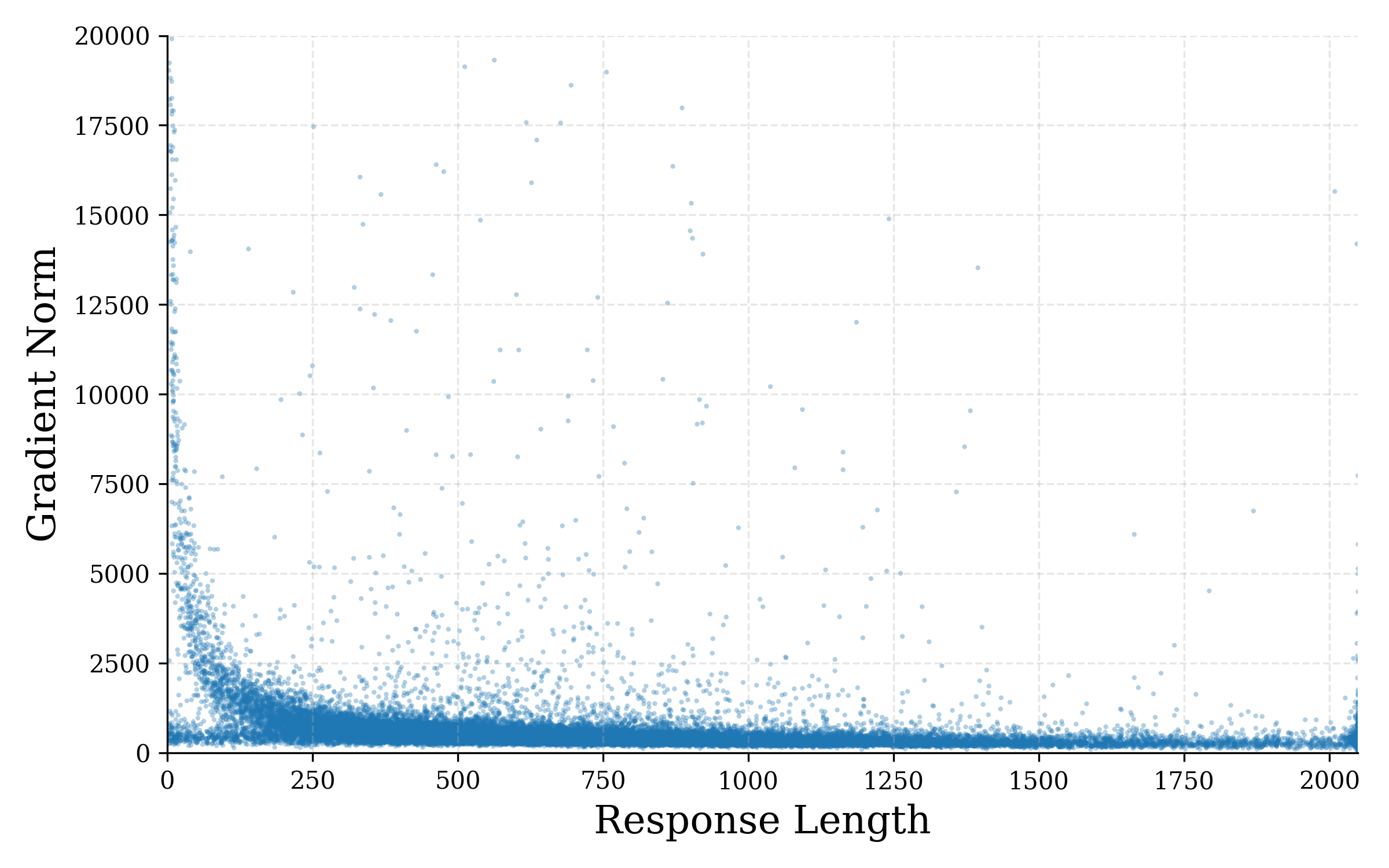}\label{fig:length1}}
	\hfil
	\subfloat[Vanilla gradients method selects shorter examples and leads to worse performance.]{\includegraphics[width=2.4in,trim=0 3 0 0,clip]{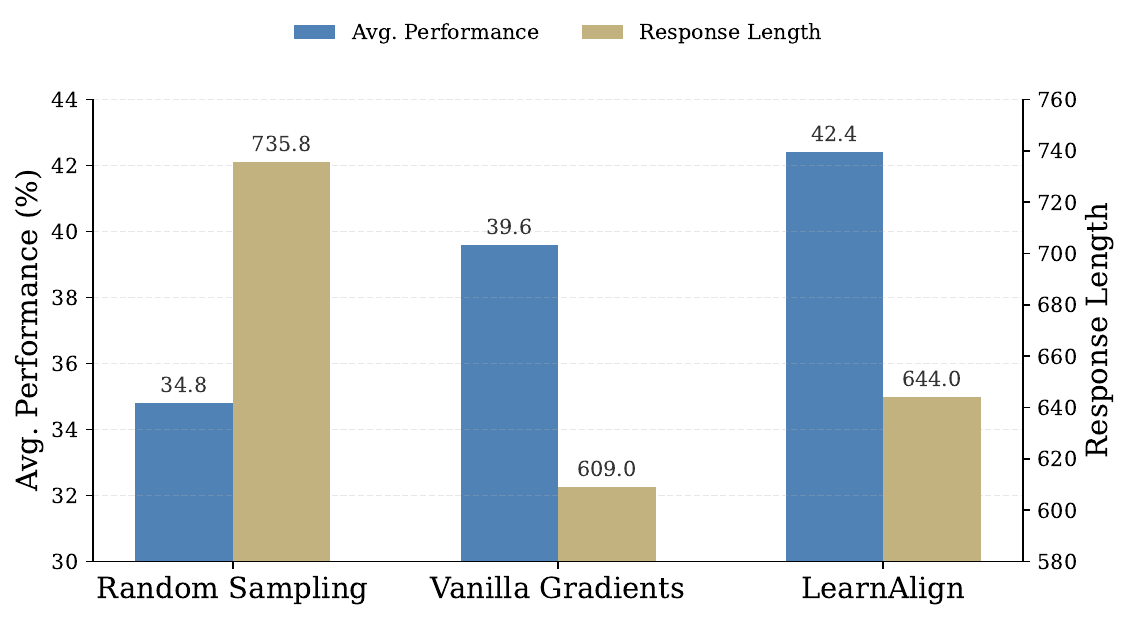}\label{fig:length2}}
	\caption{Analysis of response length and gradient-based example selection.}
\end{figure}
\subsection{Discussions}
\paragraph{\textbf{Response-length bias issue:}}

Similar to SFT, sequence-level policy gradients require averaging across tokens within a sequence. As shown in Figure~\ref{fig:length1}, the gradient norm exhibits an inverse correlation with response length, introducing a systematic bias. Consequently, as shown in Figure~\ref{fig:length2}, compared with \textit{LearnAlign}, which replaces gradient norms with success-rate–based learnability, the data selected by vanilla gradient matching yields much shorter responses and lower performance. Given that incorrect responses may lead to longer outputs, \textit{LearnAlign} selects data with more moderate response lengths between vanilla gradient and random, and achieves higher average performance. Therefore, success-rate-based learnability serves as a more suitable indicator than raw gradient norms.

\begin{table}[t]
\centering
\caption{\label{tab:ablation_study}Ablation study of our method with Qwen2.5-1.5B-Instruct and Qwen2.5-3B model.} 
\renewcommand{\arraystretch}{0.9}
\setlength{\tabcolsep}{6pt}
\resizebox{\linewidth}{!}{
\begin{tabular}{@{}c|cc|ccc@{}}
\toprule
\textbf{Model} & \multicolumn{2}{c|}{\textbf{Qwen2.5-1.5B-Instruct}} & \multicolumn{3}{c}{\textbf{Qwen2.5-3B}} \\
\midrule
\textbf{Data Size} & \textbf{1,000} & \textbf{2,000} & \textbf{1,000} & \textbf{1,000} & \textbf{1,000} \\
\textbf{Benchmark} & \textbf{GSM8K} & \textbf{GSM8K} & \textbf{GSM8K} & \textbf{MATH500} & \textbf{AMC2023} \\
\midrule
 \textbf{\textit{LearnAlign}} & \textbf{77.5} & \textbf{78.3} &  \textbf{79.3} &  \textbf{60.2}&   \textbf{28.3}\\
w/o warmup training & 76.6 & 76.6 & 76.7& 58.2&  26.1\\
w/o the data learnability & 75.6 & 76.7 & 77.5& 58.4& 28.3 \\
w/ feature similarity & 75.7 & 76.6 & 79.1& 57.6& 27.5 \\
\bottomrule
\end{tabular}}
\end{table}

\begin{table}[t]
\centering
\caption{
Comparison of time cost for training Qwen2.5-3B on the DAPO-MATH-17K selected 1,000 subsets with different methods. Time is reported in hours on a single GPU. In the DAPO-MATH-17K experiments, inspired by~\citep{lin2025cppo}, we calculate the gradient of one correct rollout for each sample. * means that we calculate the gradients of all rollouts for each sample.
}
\renewcommand{\arraystretch}{1.0}
\setlength{\tabcolsep}{6pt}
\resizebox{\linewidth}{!}{
\begin{tabular}{@{}c|cccc@{}}
\toprule
\textbf{Method} & \textbf{Data Selection Time} & \textbf{Training Time} & \textbf{Speedup} & \textbf{Avg. performance}\\
\midrule
FULL & - & 42.3h & x1.00 &44.9\\
LIMR & 42.3h & 2.4h  & x0.95 &39.0\\
\method & 8.9h & 2.4h & x3.74 &42.4\\
\method* & 22.8h & 2.4h & x1.68 & 43.3 \\
\bottomrule
\end{tabular}}
\label{time}
\end{table} 

\paragraph{\textbf{Ablation studies:}}
\label{ablation}
We conducted three ablation studies on the GSM8K dataset with 1,000 and 2,000 problems, and the DAPO-MATH-17K dataset with 1,000 problems: (1) removing the warmup phase; (2) omitting the learnability metric; and (3) replacing the cosine similarity between gradients with a feature-similarity measure~\citep{ivison2025large}. 
As shown in Table~\ref{tab:ablation_study}, the removal of any single component leads to a decline in performance. 
It indicates that the warmup phase, the learnability metric, and gradient similarities each make a significant contribution to letting the data selection method capture the model’s current capability.  These findings align with the extended results in Appendix~\ref{longer_ablation_study}, further confirming that both warmup training and data learnability play essential roles in the effectiveness of the proposed method.

\paragraph{\textbf{Time-cost analysis:}}
As shown in Table~\ref{time}, our approach offers a more practical solution for RLVR post-training, compared with the alternative RLVR data selection method LIMR~\citep{li2025limr} that requires multi-epoch training on the full dataset. The computational analysis of all steps and a detailed discussion can be found in Appendix~\ref{Detailed Discussion on time}.

\begin{table}[t]
\centering
\caption{\label{tab:longer}Performance of our method with various training steps. \textbf{*} The FULL method on Qwen2.5-3B uses 2,174 training steps, and when training on Qwen2.5-7B, it uses 1,000 training steps with a batch size of 256 to support long-time training and prevent training crash.}
\renewcommand{\arraystretch}{0.95}
\setlength{\tabcolsep}{4pt}
\resizebox{\linewidth}{!}{\begin{tabular}{@{}c|ccc|ccc@{}}
\toprule
\textbf{Method} &
\multicolumn{3}{c|}{\textbf{Qwen2.5-3B}} &
\multicolumn{3}{c}{\textbf{Qwen2.5-7B}} \\
\cmidrule(lr){2-4} \cmidrule(lr){5-7}
& \textbf{GSM8K} & \textbf{MATH500} & \textbf{AMC2023} & \textbf{GSM8K} & \textbf{MATH500} & \textbf{AMC2023} \\
\midrule
FULL* & 83.6 & 65.8 & 31.0 & 90.0 &  \textbf{77.6} & 47.3 \\
\midrule
\method~(250 steps)  & 79.3 & 60.2 & 28.3 & 88.3 & 70.4 & 35.4 \\
\method~(500 steps)  & 80.7 & 63.4 & 31.5 & 89.0 & 75.3 & 43.8 \\
\method~(1,000 steps) & 82.9 & 64.6 & 35.2 &  \textbf{90.4} & 76.7 &  \textbf{48.6} \\
\method~(2,000 steps) & \textbf{83.8} & \textbf{67.8} & \textbf{36.9} & - & - & - \\
\bottomrule
\end{tabular}}
\end{table}

\paragraph{\textbf{Training step discussion}}
To examine whether the selected subset constrains the final achievable performance, we train the \textit{LearnAlign}-selected data with steps from 250 to 2000. As shown in Table~\ref{tab:longer}, training with more steps on the selected subset reaches the full-data performance and even surpasses it.

\begin{figure}[t]
		\centering
\includegraphics[width=0.95\linewidth]{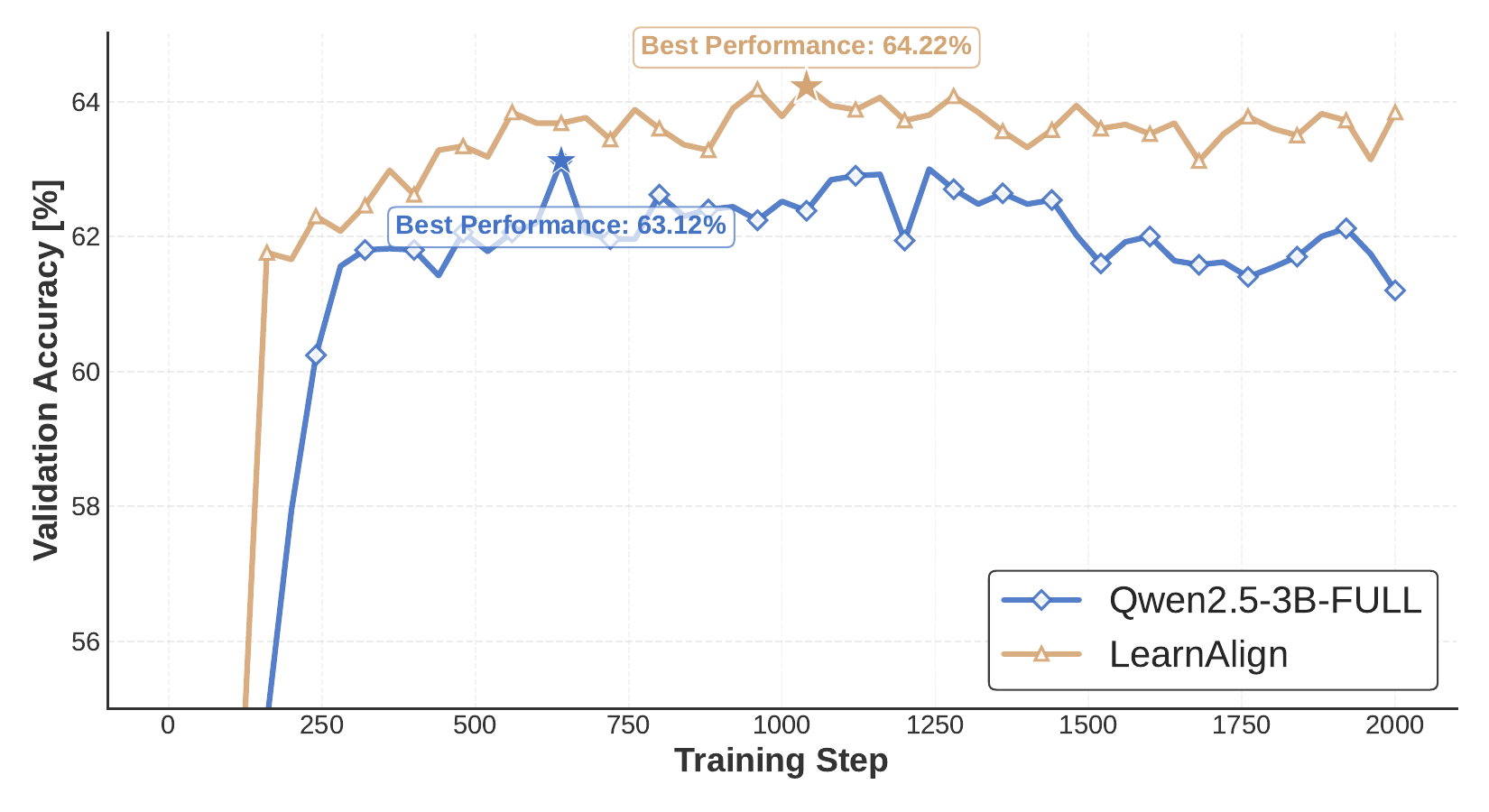}
		\caption{Validation accuracy vs. training step for \textit{LearnAlign} and the FULL method with Qwen2.5-3B. The validation is conducted on the validation set of the MATH dataset~\citep{hendrycksmath2021}. }
		\label{fig:convergence}
	\end{figure}

\paragraph{\textbf{Convergence behavior analysis}}
As shown in Figure~\ref{fig:convergence}, the full-data training peaks at 63.12\% validation accuracy at step 640, whereas \textit{LearnAlign} reaches the same accuracy by step 440, using 31\% fewer steps. This indicates substantially faster convergence under identical budgets. \textit{LearnAlign} then surpasses the full-data baseline’s peak, achieving 64.22\% at step 1040, after which its curve remains stable with a smoother plateau.

%% file: main/tabs/main_res.tex
\begin{table}[t]
\centering
\caption{\label{tab:main_results}
Comparison of data selection methods on GSM8K test set. We train {Qwen2.5-1.5B-Instruct} on the GSM8K training selected subset.}
\renewcommand{\arraystretch}{1.05}
\setlength{\tabcolsep}{8pt}
\resizebox{\linewidth}{!}{
\begin{tabular}{@{}c|cccc@{}}
\toprule
\multirow{2}{*}{\textbf{Data Selection Method}} & \multicolumn{4}{c}{\textbf{Selected Data Size}} \\
\cmidrule(lr){2-5}
 & \textbf{100} & \textbf{500} & \textbf{1,000} & \textbf{2,000}  \\ 
\midrule
Qwen2.5-1.5B-Instruct      & \multicolumn{4}{c}{55.7} \\
Qwen2.5-1.5B-Instruct-FULL & \multicolumn{4}{c}{77.0} \\
\midrule
Random Sampling        & 73.1         & 75.1        & 75.6         & 75.5           \\
PPL-Top~\citep{laurenccon2022bigscience}           & 72.5       & 75.8        & 74.6         & 75.2        \\
PPL-Middle~\citep{ankner2024perplexed}     & 72.8       & 74.7        & 75.0         & 74.2        \\
IFD~\citep{li2023quantity}            & 72.0       & 76.0        & 75.6         & 75.4   
 \\
Token Length~\citep{xia2024rethinking}            & 72.3      & 74.4       & 76.2      & 75.6
\\
SelectIT~\citep{liu2024selectit}        & 72.8       & 75.7        & 75.6         & 75.5        \\ 
\midrule
LIMR~\citep{li2025limr}        & 74.2      & 76.2       & 76.1        & 76.7     \\
\textbf{\textit{LearnAlign}}   & \textbf{74.8} & \textbf{76.4} & \textbf{77.5} & \textbf{78.3} \\ 
\bottomrule
\end{tabular}}
\vspace{-0.6cm}
\end{table}

%% file: main/tabs/main_res2.tex
\begin{table*}[t]
\centering
\caption{\label{tab:main_results2}
Comparison of data selection methods on four math benchmarks (GSM8K, MATH500, AMC2023, AIME2024) and one code benchmark (CRUX). We train {Qwen2.5-3B} and {Qwen2.5-7B} on the DAPO-MATH-17K selected subset with 1,000 data points.}
\setlength{\tabcolsep}{5pt}
\fontsize{8}{10}\selectfont
\resizebox{\linewidth}{!}{
\begin{tabular}{c|cccccc}
\toprule
 \textbf{Data Selection Method}&\textbf{GSM8K} &\textbf{MATH500}  & \textbf{AMC2023}  & \textbf{AIME2024}  & \textbf{CRUX}  & Avg.\\
 \midrule
 Qwen2.5-3B         & 20.1&52.2&8.1&3.3&14.6&19.7 \\
Qwen2.5-3B-FULL   & 83.6 &65.8&31.0&20.0&24.3&44.9\\
\midrule
Random Sampling         & 70.5&53.4&19.4&13.3& 17.4&34.8\\
PPL-Top~\citep{laurenccon2022bigscience} &71.3& 57.8 & 23.0 & 13.3 &16.1 &36.3 \\
PPL-Middle~\citep{ankner2024perplexed}      &70.1   & 54.0 &24.1&10.0    & 18.1&35.3   \\
IFD~\citep{li2023quantity}             & 70.9  & 54.4  &23.3&6.7 &15.4& 34.1  \\
Token Length~\citep{xia2024rethinking}    &35.7    &  50.4   & 18.5 &16.7 & 15.3 & 27.3 \\
SelectIT~\citep{liu2024selectit}      & 70.1   & {60.2}  &25.2 &16.7 &17.8&38.0    \\
\midrule
LIMR~\citep{li2025limr}      &74.0  & 55.6 & 25.6&\textbf{23.3}&16.5&39.0    \\
\textbf{\textit{LearnAlign}}   & \textbf{79.3}&\textbf{60.2}&\textbf{28.3}&\textbf{23.3}&\textbf{21.0} &\textbf{42.4}\\
\midrule
\midrule
Qwen2.5-7B        & 26.4&67.2&18.1&16.7&25.1&30.7\\
Qwen2.5-7B-FULL   & 89.8& 76.4&47.0&30.0&51.1&58.9 \\
\midrule
Random Sampling         & 81.1  &65.0& 30.1&23.3&  40.8 &48.1  \\
PPL-Top~\citep{laurenccon2022bigscience}        &  87.7 & 65.4 & 28.0&20.0&42.5&48.7      \\
PPL-Middle~\citep{ankner2024perplexed}      &  85.1& 64.4&27.3&16.7& 43.3&47.4       \\
IFD~\citep{li2023quantity}             & 79.4&58.6    & 29.8&13.3& 34.9&  43.2  \\
Token Length~\citep{xia2024rethinking}            &  81.4  & 62.2 &31.0&20.0&38.1&46.5    \\
SelectIT~\citep{liu2024selectit}      & 85.4&67.0&32.7    & 26.7 &41.5 &50.7   \\
\midrule
LIMR~\citep{li2025limr}      & 84.2   &  61.6&27.1&16.7& 39.9&45.9   \\
\textbf{\textit{LearnAlign}}   & \textbf{88.3}&\textbf{70.4}&\textbf{35.4}&\textbf{30.0}&\textbf{44.0}&\textbf{54.6 } \\
\bottomrule
\end{tabular}}
\end{table*}

%% file: custom.bib
@article{johnson1984extensions,
  title={Extensions of Lipschitz mappings into a Hilbert space},
  author={Johnson, William B and Lindenstrauss, Joram and others},
  journal={Contemporary mathematics},
  volume={26},
  number={189-206},
  pages={1},
  year={1984}
}

@inproceedings{yeh2022first,
  title={First is better than last for language data influence},
  author={Yeh, Chih-Kuan and Taly, Ankur and Sundararajan, Mukund and Liu, Frederick and Ravikumar, Pradeep},
  booktitle={NeurIPS},
  volume={35},
  pages={32285--32298},
  year={2022}
}

@inproceedings{hu2022lora,
  title={Lora: Low-rank adaptation of large language models.},
  author={Hu, Edward J and Shen, Yelong and Wallis, Phillip and Allen-Zhu, Zeyuan and Li, Yuanzhi and Wang, Shean and Wang, Lu and Chen, Weizhu and others},
  booktitle={ICLR},
  year={2022}
}

@inproceedings{williams2000using,
  title={Using the Nystr{\"o}m method to speed up kernel machines},
  author={Williams, Christopher and Seeger, Matthias},
  booktitle={NeurIPS},
  volume={13},
  year={2000}
}

@inproceedings{gong2025two,
  title={A two-stage data selection framework for data-efficient model training on edge devices},
  author={Gong, Chen and Xing, Rui and Zheng, Zhenzhe and Wu, Fan},
  booktitle={KDD},
  pages={673--684},
  year={2025}
}

@inproceedings{agarwalneural,
  title={Neural Networks for Learnable and Scalable Influence Estimation of Instruction Fine-Tuning Data},
  author={Agarwal, Ishika and Hakkani-T{\"u}r, Dilek},
  booktitle={NeurIPS},
  year={2025}
}

@article{shao2024deepseekmath,
  title={Deepseekmath: Pushing the limits of mathematical reasoning in open language models},
  author={Shao, Zhihong and Wang, Peiyi and Zhu, Qihao and Xu, Runxin and Song, Junxiao and Bi, Xiao and Zhang, Haowei and Zhang, Mingchuan and Li, YK and Wu, Y and others},
  journal={arXiv preprint arXiv:2402.03300},
  year={2024}
}

@inproceedings{xia2024less,
  title={LESS: selecting influential data for targeted instruction tuning},
  author={Xia, Mengzhou and Malladi, Sadhika and Gururangan, Suchin and Arora, Sanjeev and Chen, Danqi},
  booktitle={ICML},
  pages={54104--54132},
  year={2024}
}

@article{liu2025mtp,
  title={L-MTP: Leap Multi-Token Prediction Beyond Adjacent Context for Large Language Models},
  author={Liu, Xiaohao and Xia, Xiaobo and Zhao, Weixiang and Zhang, Manyi and Yu, Xianzhi and Su, Xiu and Yang, Shuo and Ng, See-Kiong and Chua, Tat-Seng},
  journal={arXiv preprint arXiv:2505.17505},
  year={2025}
}

@article{luo2024mmevol,
  title={Mmevol: Empowering multimodal large language models with evol-instruct},
  author={Luo, Run and Zhang, Haonan and Chen, Longze and Lin, Ting-En and Liu, Xiong and Wu, Yuchuan and Yang, Min and Wang, Minzheng and Zeng, Pengpeng and Gao, Lianli and others},
  journal={arXiv preprint arXiv:2409.05840},
  year={2024}
}

@article{luo2025gui,
  title={Gui-r1: A generalist r1-style vision-language action model for gui agents},
  author={Luo, Run and Wang, Lu and He, Wanwei and Xia, Xiaobo},
  journal={arXiv preprint arXiv:2504.10458},
  year={2025}
}

@article{li2025limr,
  title={Limr: Less is more for rl scaling},
  author={Li, Xuefeng and Zou, Haoyang and Liu, Pengfei},
  journal={arXiv preprint arXiv:2502.11886},
  year={2025}
}

@inproceedings{zhou2023lima,
  title={Lima: Less is more for alignment},
  author={Zhou, Chunting and Liu, Pengfei and Xu, Puxin and Iyer, Srinivasan and Sun, Jiao and Mao, Yuning and Ma, Xuezhe and Efrat, Avia and Yu, Ping and Yu, Lili and others},
  booktitle={NeurIPS},
  volume={36},
  pages={55006--55021},
  year={2023}
}

@article{lin2025cppo,
  title={Cppo: Accelerating the training of group relative policy optimization-based reasoning models},
  author={Lin, Zhihang and Lin, Mingbao and Xie, Yuan and Ji, Rongrong},
  journal={arXiv preprint arXiv:2503.22342},
  year={2025}
}

@book{sutton1998reinforcement,
  title={Reinforcement learning: An introduction},
  author={Sutton, Richard S and Barto, Andrew G and others},
  volume={1},
  number={1},
  year={1998},
  publisher={MIT press Cambridge}
}

@article{foster2025learning,
  title={Learning to Reason at the Frontier of Learnability},
  author={Foster, Thomas and Foerster, Jakob},
  journal={arXiv preprint arXiv:2502.12272},
  year={2025}
}

@article{tzannetos2023proximal,
  title={Proximal curriculum for reinforcement learning agents},
  author={Tzannetos, Georgios and Ribeiro, B{\'a}rbara Gomes and Kamalaruban, Parameswaran and Singla, Adish},
  journal={arXiv preprint arXiv:2304.12877},
  year={2023}
}

@inproceedings{florensa2018automatic,
  title={Automatic goal generation for reinforcement learning agents},
  author={Florensa, Carlos and Held, David and Geng, Xinyang and Abbeel, Pieter},
  booktitle={ICML},
  pages={1515--1528},
  year={2018},
}

@article{razin2025makes,
  title={What makes a reward model a good teacher? an optimization perspective},
  author={Razin, Noam and Wang, Zixuan and Strauss, Hubert and Wei, Stanley and Lee, Jason D and Arora, Sanjeev},
  journal={arXiv preprint arXiv:2503.15477},
  year={2025}
}

@article{jaech2024openai,
  title={Openai o1 system card},
  author={Jaech, Aaron and Kalai, Adam and Lerer, Adam and Richardson, Adam and El-Kishky, Ahmed and Low, Aiden and Helyar, Alec and Madry, Aleksander and Beutel, Alex and Carney, Alex and others},
  journal={arXiv preprint arXiv:2412.16720},
  year={2024}
}

@article{guo2025deepseek,
  title={Deepseek-r1: Incentivizing reasoning capability in llms via reinforcement learning},
  author={Guo, Daya and Yang, Dejian and Zhang, Haowei and Song, Junxiao and Zhang, Ruoyu and Xu, Runxin and Zhu, Qihao and Ma, Shirong and Wang, Peiyi and Bi, Xiao and others},
  journal={arXiv preprint arXiv:2501.12948},
  year={2025}
}

@article{team2025kimi,
  title={Kimi k1. 5: Scaling reinforcement learning with llms},
  author={Team, Kimi and Du, Angang and Gao, Bofei and Xing, Bowei and Jiang, Changjiu and Chen, Cheng and Li, Cheng and Xiao, Chenjun and Du, Chenzhuang and Liao, Chonghua and others},
  journal={arXiv preprint arXiv:2501.12599},
  year={2025}
}

@article{ye2025limo,
  title={LIMO: Less is More for Reasoning},
  author={Ye, Yixin and Huang, Zhen and Xiao, Yang and Chern, Ethan and Xia, Shijie and Liu, Pengfei},
  journal={arXiv preprint arXiv:2502.03387},
  year={2025}
}

@article{li2023quantity,
  title={From quantity to quality: Boosting llm performance with self-guided data selection for instruction tuning},
  author={Li, Ming and Zhang, Yong and Li, Zhitao and Chen, Jiuhai and Chen, Lichang and Cheng, Ning and Wang, Jianzong and Zhou, Tianyi and Xiao, Jing},
  journal={arXiv preprint arXiv:2308.12032},
  year={2023}
}

@article{ankner2024perplexed,
  title={Perplexed by perplexity: Perplexity-based data pruning with small reference models},
  author={Ankner, Zachary and Blakeney, Cody and Sreenivasan, Kartik and Marion, Max and Leavitt, Matthew L and Paul, Mansheej},
  journal={arXiv preprint arXiv:2405.20541},
  year={2024}
}

@article{wang2025reinforcement,
  title={Reinforcement Learning for Reasoning in Large Language Models with One Training Example},
  author={Wang, Yiping and Yang, Qing and Zeng, Zhiyuan and Ren, Liliang and Liu, Lucas and Peng, Baolin and Cheng, Hao and He, Xuehai and Wang, Kuan and Gao, Jianfeng and others},
  journal={arXiv preprint arXiv:2504.20571},
  year={2025}
}

@inproceedings{hendrycksmath2021,
  title={Measuring Mathematical Problem Solving With the MATH Dataset},
  author={Dan Hendrycks and Collin Burns and Saurav Kadavath and Akul Arora and Steven Basart and Eric Tang and Dawn Song and Jacob Steinhardt},
  booktitle={NeurIPS},
  year={2021}
}

@article{bae2025online,
  title={Online difficulty filtering for reasoning oriented reinforcement learning},
  author={Bae, Sanghwan and Hong, Jiwoo and Lee, Min Young and Kim, Hanbyul and Nam, JeongYeon and Kwak, Donghyun},
  journal={arXiv preprint arXiv:2504.03380},
  year={2025}
}

@article{liu2025understanding,
  title={Understanding r1-zero-like training: A critical perspective},
  author={Liu, Zichen and Chen, Changyu and Li, Wenjun and Qi, Penghui and Pang, Tianyu and Du, Chao and Lee, Wee Sun and Lin, Min},
  journal={arXiv preprint arXiv:2503.20783},
  year={2025}
}

@inproceedings{pruthi2020estimating,
  title={Estimating training data influence by tracing gradient descent},
  author={Pruthi, Garima and Liu, Frederick and Kale, Satyen and Sundararajan, Mukund},
  booktitle={NeurIPS},
  volume={33},
  pages={19920--19930},
  year={2020}
}

@article{schulman2017proximal,
  title={Proximal policy optimization algorithms},
  author={Schulman, John and Wolski, Filip and Dhariwal, Prafulla and Radford, Alec and Klimov, Oleg},
  journal={arXiv preprint arXiv:1707.06347},
  year={2017}
}

@article{liu2024less,
  title={Less is More: High-value Data Selection for Visual Instruction Tuning},
  author={Liu, Zikang and Zhou, Kun and Zhao, Wayne Xin and Gao, Dawei and Li, Yaliang and Wen, Ji-Rong},
  journal={arXiv preprint arXiv:2403.09559},
  year={2024}
}

@inproceedings{wang2020optimizing,
  title={Optimizing data usage via differentiable rewards},
  author={Wang, Xinyi and Pham, Hieu and Michel, Paul and Anastasopoulos, Antonios and Carbonell, Jaime and Neubig, Graham},
  booktitle={ICML},
  pages={9983--9995},
  year={2020}
}

@article{mackay1992information,
  title={Information-based objective functions for active data selection},
  author={MacKay, David JC},
  journal={Neural computation},
  volume={4},
  number={4},
  pages={590--604},
  year={1992},
  publisher={MIT Press One Rogers Street, Cambridge, MA 02142-1209, USA journals-info~…}
}

@article{chaiklin2003zone,
  title={The zone of proximal development in Vygotsky’s analysis of learning and instruction},
  author={Chaiklin, Seth and others},
  journal={Vygotsky’s educational theory in cultural context},
  volume={1},
  number={2},
  pages={39--64},
  year={2003},
  publisher={Cambridge, UK: Cambridge University Press.}
}

@inproceedings{laurenccon2022bigscience,
  title={The bigscience roots corpus: A 1.6 tb composite multilingual dataset},
  author={Lauren{\c{c}}on, Hugo and Saulnier, Lucile and Wang, Thomas and Akiki, Christopher and Villanova del Moral, Albert and Le Scao, Teven and Von Werra, Leandro and Mou, Chenghao and Gonz{\'a}lez Ponferrada, Eduardo and Nguyen, Huu and others},
  booktitle={NeurIPS},
  volume={35},
  pages={31809--31826},
  year={2022}
}

@article{AMC2023mathematical,
  title={American mathematics competitions},
  journal={The Mathematical Association of America},
  year={2023}
}

@article{AIME2024mathematical,
  title={Maa invitational competitions- American Invitational Mathe
matics Examination},
  journal={The Mathematical Association of America},
  year={2024}
}

@article{hendrycks2021measuring,
  title={Measuring mathematical problem solving with the math dataset},
  author={Hendrycks, Dan and Burns, Collin and Kadavath, Saurav and Arora, Akul and Basart, Steven and Tang, Eric and Song, Dawn and Steinhardt, Jacob},
  journal={arXiv preprint arXiv:2103.03874},
  year={2021}
}

@article{cobbe2021training,
  title={Training verifiers to solve math word problems},
  author={Cobbe, Karl and Kosaraju, Vineet and Bavarian, Mohammad and Chen, Mark and Jun, Heewoo and Kaiser, Lukasz and Plappert, Matthias and Tworek, Jerry and Hilton, Jacob and Nakano, Reiichiro and others},
  journal={arXiv preprint arXiv:2110.14168},
  year={2021}
}

@article{xia2024rethinking,
  title={Rethinking data selection at scale: Random selection is almost all you need},
  author={Xia, Tingyu and Yu, Bowen and Dang, Kai and Yang, An and Wu, Yuan and Tian, Yuan and Chang, Yi and Lin, Junyang},
  journal={arXiv preprint arXiv:2410.09335},
  year={2024}
}

@article{ivison2025large,
  title={Large-Scale Data Selection for Instruction Tuning},
  author={Ivison, Hamish and Zhang, Muru and Brahman, Faeze and Koh, Pang Wei and Dasigi, Pradeep},
  journal={arXiv preprint arXiv:2503.01807},
  year={2025}
}

@inproceedings{lu2023instag,
  title={Instag: Instruction tagging for analyzing supervised fine-tuning of large language models},
  author={Lu, Keming and Yuan, Hongyi and Yuan, Zheng and Lin, Runji and Lin, Junyang and Tan, Chuanqi and Zhou, Chang and Zhou, Jingren},
  booktitle={ICLR},
  year={2023}
}

@article{chen2023alpagasus,
  title={Alpagasus: Training a better alpaca with fewer data},
  author={Chen, Lichang and Li, Shiyang and Yan, Jun and Wang, Hai and Gunaratna, Kalpa and Yadav, Vikas and Tang, Zheng and Srinivasan, Vijay and Zhou, Tianyi and Huang, Heng and others},
  journal={arXiv preprint arXiv:2307.08701},
  year={2023}
}

@article{liu2024selectit,
  title={SelectIT: Selective Instruction Tuning for Large Language Models via Uncertainty-Aware Self-Reflection},
  author={Liu, Liangxin and Liu, Xuebo and Wong, Derek F and Li, Dongfang and Wang, Ziyi and Hu, Baotian and Zhang, Min},
  journal={arXiv preprint arXiv:2402.16705},
  year={2024}
}

@article{li2023one,
  title={One Shot Learning as Instruction Data Prospector for Large Language Models},
  author={Li, Yunshui and Hui, Binyuan and Xia, Xiaobo and Yang, Jiaxi and Yang, Min and Zhang, Lei and Si, Shuzheng and Liu, Junhao and Liu, Tongliang and Huang, Fei and others},
  journal={arXiv preprint arXiv:2312.10302},
  year={2023}
}

@inproceedings{zhang2024m3d,
  title={M3d: Dataset condensation by minimizing maximum mean discrepancy},
  author={Zhang, Hansong and Li, Shikun and Wang, Pengju and Zeng, Dan and Ge, Shiming},
  booktitle={AAAI},
  volume={38},
  number={8},
  pages={9314--9322},
  year={2024}
}

@inproceedings{zhang2024dance,
  title={DANCE: Dual-View Distribution Alignment for Dataset Condensation},
  author={Zhang, Hansong and Li, Shikun and Lin, Fanzhao and Wang, Weiping and Qian, Zhenxing and Ge, Shiming},
  booktitle={IJCAI},
  year={2024}
}

@article{gu2024cruxeval,
  title={Cruxeval: A benchmark for code reasoning, understanding and execution},
  author={Gu, Alex and Rozi{\`e}re, Baptiste and Leather, Hugh and Solar-Lezama, Armando and Synnaeve, Gabriel and Wang, Sida I},
  journal={arXiv preprint arXiv:2401.03065},
  year={2024}
}

@article{yu2025dapo,
  title={Dapo: An open-source llm reinforcement learning system at scale},
  author={Yu, Qiying and Zhang, Zheng and Zhu, Ruofei and Yuan, Yufeng and Zuo, Xiaochen and Yue, Yu and Dai, Weinan and Fan, Tiantian and Liu, Gaohong and Liu, Lingjun and others},
  journal={arXiv preprint arXiv:2503.14476},
  year={2025}
}

@inproceedings{li2022estimating,
  title={Estimating noise transition matrix with label correlations for noisy multi-label learning},
  author={Li, Shikun and Xia, Xiaobo and Zhang, Hansong and Zhan, Yibing and Ge, Shiming and Liu, Tongliang},
  booktitle={NeurIPS},
  volume={35},
  pages={24184--24198},
  year={2022}
}

@inproceedings{li2022selective,
  title={Selective-supervised contrastive learning with noisy labels},
  author={Li, Shikun and Xia, Xiaobo and Ge, Shiming and Liu, Tongliang},
  booktitle={CVPR},
  pages={316--325},
  year={2022}
}
